%% file: preprint.tex
\documentclass[11pt]{article}

\usepackage[preprint]{acl}

\usepackage{times}
\usepackage{latexsym}
\usepackage[ruled,vlined]{algorithm2e}
\usepackage[T1]{fontenc}
\usepackage[most]{tcolorbox}
\usepackage[utf8]{inputenc}
\usepackage{subcaption}
\usepackage{microtype}
\usepackage{cuted}
\usepackage{inconsolata}
\usepackage{amsmath}
\usepackage{mathtools}
\usepackage{amssymb}
\usepackage{xspace}
\usepackage[most]{tcolorbox}
\usepackage{xcolor}
\usepackage[table]{xcolor}
\usepackage{enumitem}
\newtcolorbox{promptbox}{
  colback=gray!4,
  colframe=gray!55,
  boxrule=0.5pt,
  arc=2pt,
  left=6pt,
  right=6pt,
  top=6pt,
  bottom=6pt,
  fontupper=\ttfamily\footnotesize,
  breakable
}
\usepackage{booktabs}
\usepackage{graphicx}

\usepackage{hyperref}
\usepackage{cleveref}

\title{OGLS-SD: On-Policy Self-Distillation with Outcome-Guided Logit Steering for LLM Reasoning}

\author{
Yuxiao Yang \\ UNC Chapel Hill\\ \texttt{yxyang@unc.edu}
\And
Xiaoyun Wang \\ NVIDIA\\ \texttt{xiaoyunw@nvidia.com}
\And
Weitong Zhang\\ UNC Chapel Hill\\ \texttt{weitongz@unc.edu}
}
\newcommand{\methodname}{OGLS-SD\xspace}

\begin{document}

\maketitle
\begin{abstract}
We study \textbf{on-policy self-distillation} (OPSD), where a language model improves its reasoning ability by distilling privileged teacher distributions along its own on-policy trajectories. Despite its promise, OPSD can suffer from training instability due to a pattern mismatch between teacher and student responses. Self-reflected teacher responses may introduce reflection-induced biases and response templates that miscalibrate token-level supervision, ultimately harming the student's reasoning ability. To mitigate this issue, we propose \methodname, an outcome-guided logit-steering framework that leverages verifiable outcome rewards to calibrate privileged teacher logits. Specifically, \methodname contrasts teacher logits induced by successful and failed on-policy trajectories, constructing an outcome-discriminative steering direction for token-level guidance. Experiments on mathematical reasoning benchmarks show that \methodname stabilizes self-distillation and improves performance over standard OPSD and other variants.
\end{abstract}
\input{latex/1.Intro_modify}
\input{latex/2.relatedworks}
\input{latex/3.Preliminaries}
\input{latex/4.Methods}
\input{latex/5.Experiments}
\input{latex/6.Conclusion}

\bibliography{custom}
\clearpage
\appendix
\input{latex/7.Appendix}

\end{document}

%% file: latex/1.Intro_modify.tex
\section{Introduction}
Reinforcement learning with verifiable rewards (RLVR; \citealt{grpo,flowrl,gspo}) provides an effective paradigm for improving LLM reasoning during post-training, but its supervision is typically sparse and sequence-level: all tokens in a sampled trajectory are optimized according to the same outcome reward. This makes fine-grained token-level credit assignment
difficult and can lead to inefficient learning~\citep{opsd}.
To obtain denser supervision, \textbf{on-policy distillation} (OPD; \citealt{opd1}) has recently emerged as a promising alternative. In OPD, the
student leverages its own on-policy generations to construct token-level guidance, leading to strong empirical performance in recent reasoning post-training
studies~\citep{opd1,opd2,opd3,opd4,rethinkopd}. OPD-style training has also been adopted or explored in industrial pipelines such as MiMo~\citep{mimo}, Qwen3~\citep{qwen3}, GLM5~\citep{glm5}, and DeepSeek-V4~\citep{deepseekv4}.

Motivated by the success of OPD, a growing line of work has leveraged LLMs' self-reflection~\citep{reflection} and self-judgment~\citep{llmjudge} abilities to enable self-improvement without relying on a larger teacher model. This setting, referred to as \textbf{on-policy self-distillation} (OPSD), removes external teacher guidance and thus enables a more direct comparison with RLVR. In OPSD, the student model itself serves as the teacher, typically by conditioning on external privileged information during training to provide dense token-level guidance~\citep{opsd,sdpo}.

Although OPSD can improve both performance and training efficiency, recent studies have reported that vanilla OPSD may suffer from training collapse and instability~\citep{rlsd,whyopsdfail}. These observations have motivated hybrid approaches that integrate OPSD with RLVR~\citep{ding2026hdpo,unify-grpo-sdpo}. However, existing hybrid methods mainly use outcome-level rewards to regularize or optimize the student policy, while leaving the dense privileged teacher distribution itself largely unchanged. Less attention has been paid to how this privileged token-level guidance should be calibrated before being distilled into the student. This naturally raises the following question:
\begin{center}
\emph{How can we calibrate privileged teacher signals for stable on-policy self-distillation?}
\end{center}

\begin{figure*}[t]
\centering
\includegraphics[width=\linewidth]{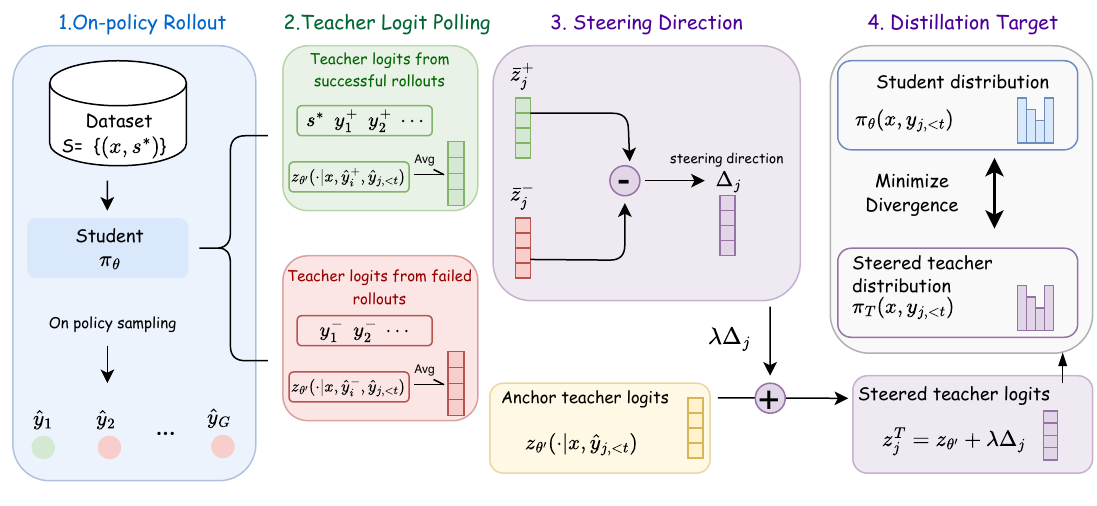}
\caption{
Illustration of \methodname.
The student model first generates $G$ on-policy rollouts for each input,
which are then verified and partitioned into correct and incorrect pools.
We compute the average teacher logits conditioned on each pool, construct
the steering direction $\Delta$ by contrasting the two averaged logits,
and add it to the anchor teacher logits to obtain the steered teacher
distribution for token-level guidance.
}
\label{fig:methodfig}
\end{figure*}
We approach this question by first examining why privileged teacher signals can become unstable distillation targets. We hypothesize that an important source of OPSD instability is a pattern mismatch between the privileged teacher and the student response. This mismatch arises from two sources. First, the teacher often constructs its response by self-reflecting on a single reference solution or successful trajectory. The resulting supervision can therefore become concentrated around a narrow output pattern, even though multiple valid reasoning paths may exist for the same problem~\citep{flowrl}. Such solution-specific bias can potentially erode the student's original capabilities~\citep{rlsd}. Second, privileged conditioning and self-reflection can introduce reasoning-pattern shifts that suppress self-checking behavior, causing the student to imitate miscalibrated token-level targets~\citep{whyopsdfail}.
We further examine this reasoning-pattern shift through the diagnostic experiment in \Cref{fig:logprob_ill}.

Motivated by these observations, we propose \methodname, \textbf{Outcome-Guided Logit Steering for Self-Distillation}, to calibrate privileged teacher signals before distillation. To mitigate solution-specific bias, \methodname samples a group of on-policy trajectories for each prompt, verifies their correctness, and aggregates the teacher logits induced by successful and failed rollouts rather than relying on a single privileged solution. To address reasoning-pattern mismatch, \methodname constructs an outcome-guided steering direction by contrasting the successful and failed guidance signals and uses the resulting
steered logits to define the teacher distribution for self-distillation. \Cref{fig:steering-vector} provides a geometric illustration of the steering intuition, and the overall framework is shown in \Cref{fig:methodfig}. In summary, our contributions are threefold:
\begin{itemize}[leftmargin=*, noitemsep]

\item We investigate the pattern mismatch between teacher--student responses in on-policy self-distillation, including the solution-specific bias and privileged-context-induced teacher pattern shifts. Through diagnostic analysis, we provide evidence that such a mismatch can make privileged guidance a miscalibrated distillation target for vanilla OPSD, which can further harm the reasoning ability of the student language model.

\item We propose \methodname, an outcome-guided self-distillation framework that calibrates privileged teacher signals by aggregating and contrasting guidance
from correct and incorrect on-policy rollouts. \methodname mitigates this mismatch and subsequently improves the reasoning ability.

\item Empirically, we show that \methodname improves both stability and performance over standard OPSD and naive integrations of RLVR objectives. These results suggest that outcome-guided calibration of privileged teacher signals enables more effective self-improvement of language models.

\end{itemize}

%% file: latex/2.relatedworks.tex
\begin{figure}[hbt]
    \centering
    \includegraphics[width=\linewidth]{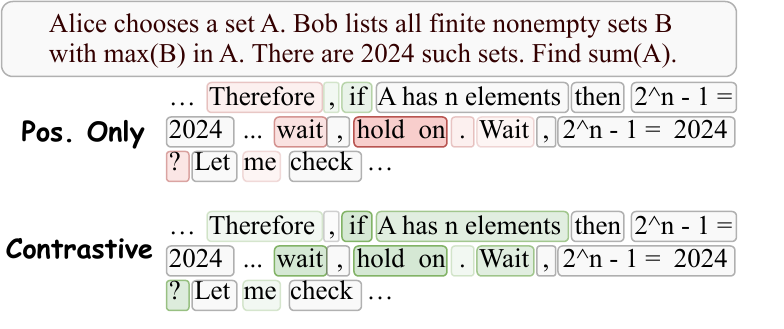}
\caption{
Reference-model diagnostic on AIME24 using Qwen3-1.7B. Colors show the guidance-induced log-probability shift
$
\Delta(y_t)=
\log \pi_T(y_t\mid x,y_{<t})
-
\log \pi_{\mathrm{ref}}(y_t\mid x,y_{<t})
$,
where red/green indicate lower/higher probability under the guided teacher and darker colors indicate larger shifts. Positive-privileged guidance suppresses intermediate reasoning tokens, while contrastive guidance mitigates this effect.
}
    \label{fig:logprob_ill}
\end{figure}

\section{Related Works}

\subsection{On-policy Distillation}

On-policy distillation (OPD) samples rollouts from the student policy and uses teacher logits computed along these on-policy trajectories to supervise the student's next-token distributions. Compared with offline distillation, OPD provides dense supervision on the states actually visited by the student, and often leads to faster and more effective performance improvement~\citep{opd1,opd2,opd3,Speculativeopd,tsd-kd}. It has also been widely adopted in industrial post-training pipelines, where it can reduce the reliance on, or even partially replace, RLVR-based post-training~\citep{qwen3,mimo}. Recent OPD variants further explore black-box teacher guidance~\citep{blackobxopd}, multi-model distillation~\citep{piflow}, and agent-level policy
distillation~\citep{refinedpolicydistillation}.
\subsection{LLM Self-Distillation}

Despite its promise, OPD typically requires a more capable teacher model, often a larger model, which may be expensive or unavailable in practice. On-policy self-distillation (OPSD) addresses this limitation by using the model itself as the teacher, while providing it with external privileged information such as a reference solution or final answer~\citep{opsd,sdpo}.

However, vanilla OPSD has been reported to suffer from unstable training and degradation in reasoning performance~\citep{whyopsdfail,rlsd}. Prior
work attributes this behavior to different forms of teacher-student mismatch. \citet{whyopsdfail} observes that self-distillation can suppress epistemic verbalization, leading to more compressed and overconfident reasoning patterns. From an information-theoretic perspective, \citet{rlsd} shows that matching a teacher conditioned on a single privileged solution is ill-posed for a student that cannot access this
solution, and that the population optimum corresponds to a marginal privileged teacher.

Combining self-distillation with RLVR has recently emerged as a promising
direction~\citep{rlsd,unify-grpo-sdpo,sdzero,luffy,aligndistill}. Existing
methods typically stabilize self-distillation by adding outcome-level RL objectives or routing samples between RL and distillation losses. For example, prior work uses self-distillation to provide denser credit assignment for GRPO~\citep{rlsd}, or combines GRPO-style policy optimization with self-distillation objectives~\citep{unify-grpo-sdpo,ding2026hdpo}. While
effective, these approaches leave the construction of the dense privileged teacher target largely unchanged, and therefore do not directly address the teacher--student mismatch induced by privileged prompting. Our work instead calibrates the token-level teacher signal itself by constructing an outcome-contrasted logit-steering direction from correct and incorrect on-policy rollouts.

\begin{figure}[t]
    \centering
    \includegraphics[width=\linewidth]{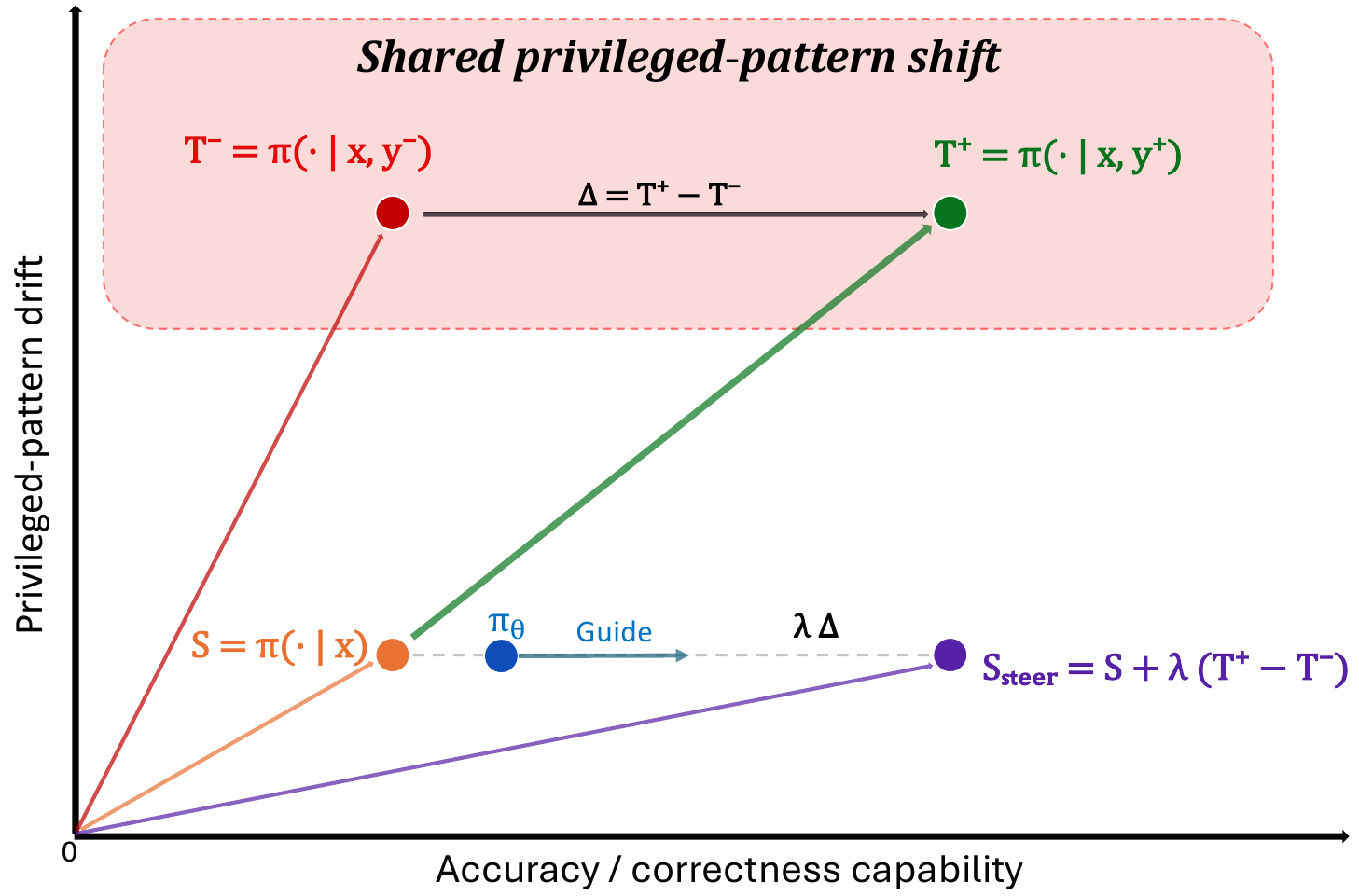}
 \caption{Illustration of outcome-guided steering. Positive and negative privileged teachers may share response-pattern shifts induced by privileged conditioning, while differing in outcome-discriminative guidance. Contrasting them attenuates shared privileged-pattern mismatch and preserves correctness-related signal.
}
    \label{fig:steering-vector}
\end{figure}

\subsection{Steering LLM Generation}

Inspired by the observation that semantic and behavioral attributes can
often be represented as approximately linear directions in model
representations~\citep{word2vec}, steering methods have been widely studied
as lightweight mechanisms for controlling LLM behavior~\citep{steerLlama,
steerllm,zhao2025mitigating}. Existing methods typically construct steering
directions from contrastive examples or learned representations, and apply
them to hidden states or output distributions to guide generation toward
desired behaviors. Most prior work focuses on controllable generation,
persona editing, safety alignment, refusal control, and hallucination
mitigation~\citep{zhao2025mitigating,learn2Steer,refusalllm}.

In contrast, using distributional steering or objective-level extrapolation as dense training-time guidance for on-policy distillation remains relatively underexplored. The closest work is EXOPD~\citep{exopd}, which introduces a reference policy $\pi_{\mathrm{ref}}$ to derive an extrapolated OPD objective in the standard OPD setting. Our work instead studies OPSD, where privileged prompting induces additional privileged-information mismatch; here, steering is used not only to amplify the teacher signal, but also to mitigate this mismatch.

%% file: latex/3.Preliminaries.tex
\section{Preliminaries}

We first introduce the notation used throughout this paper. Let $x$ denote a question prompt, and let $y$ denote a response generated by the student policy, with length $|y|$. We use $y^+$ and $y^-$ to denote correct and incorrect rollouts, respectively, as determined by a verifiable reward signal. Following the OPSD setting, each
prompt is additionally associated with privileged information $s$, such as an expert solution or reference reasoning trace, which is available to the teacher during training but not to the student at test time. We write the training dataset as $\mathcal{D}=\{(x_i,s_i)\}_{i=1}^N$, while student rollouts are sampled on policy during training.

\subsection{On-Policy Self-Distillation}
\label{opsd_disc}

For each question $x_i$, the student model first generates an on-policy response $y_i \sim \pi_{\theta}(\cdot \mid x_i)$. A teacher model with parameters $\theta'$, typically instantiated as a fixed initial model or an EMA copy of the student \citep{opsd,sdpo}, is then conditioned on additional privileged information $s_i$. The per-trajectory training objective can be written as 
\begin{equation}
\begin{aligned}
\mathcal{L}_{\mathrm{OPSD}}(\theta)
&=
\frac{1}{|y_i|}
\sum_{t=1}^{|y_i|}
D\Big(
    \mathrm{sg}\big[
        \pi_{\theta'}(\cdot \mid x_i, s_i, y_{i,<t})
    \big]
\\
&\qquad\qquad
,\pi_{\theta}(\cdot \mid x_i, y_{i,<t})
\Big), 
\end{aligned}
\label{eq:opsd-loss}
\end{equation}
where $\mathrm{sg}[\cdot]$ denotes the stop-gradient operation. We write
$D(p_T,p_\theta)$ for a token-level distribution-matching loss from teacher to student, such as KL divergence.

Intuitively, the privileged teacher is expected to provide fine-grained token-level guidance along the student's on-policy trajectory, replacing sparse sequence-level rewards with dense distributional supervision.

However, the student policy $\pi_{\theta}(\cdot \mid x,y_{<t})$ cannot access the privileged solution $s$ at test time. Directly matching the privileged teacher $\pi_{\theta'}(\cdot\mid x,s,y_{<t})$ therefore creates a teacher-student mismatch. As pointed out by~\citep{rlsd}, the corresponding population optimum within the non-privileged student function class is the marginal privileged teacher distribution
\begin{equation}
\bar P_T(\cdot\mid x,y_{<t})
=
\mathbb{E}_{s\sim P(s\mid x,y_{<t})}
\left[
\pi_{\theta'}(\cdot\mid x,s,y_{<t})
\right].
    \notag
\end{equation}
This suggests that instance-wise matching to a single privileged solution is not the ideal target for a non-privileged student.

The marginal privileged teacher provides a useful ideal target for reducing solution-specific information asymmetry. However, practical implementations can only approximate this marginal using a small number of privileged traces. These finite-sample approximations may still retain artifacts introduced by privileged prompting, such as sharper next-token distributions, reduced support
for exploratory tokens, or shifted reasoning styles~\citep{whyopsdfail}.

\subsection{Guidance in Large Language Models}

Let $c_t=(x,y_{<t})$ denote the generation context and $v$ denote a candidate next token. Contrastive steering can be viewed as modifying a base next-token distribution with a likelihood-ratio correction:
\begin{equation}
p_{\gamma}(v\mid c_t)
\propto
p_0(v\mid c_t)
\left(
\frac{p^{+}(v\mid c_t)}
{p^{-}(v\mid c_t)}
\right)^{\gamma}, \label{eq:1} 
\end{equation}
where $p_0$ is the base distribution, $p^{+}$ and $p^{-}$ correspond to positive and negative conditions, and $\gamma\ge0$ controls the steering strength. Taking the logarithm of Eq.~\eqref{eq:1} yields
\begin{align}
    \log p_{\gamma}(v \mid c_t) &\propto \log p_0(v \mid c_t) \notag\\
    & \, + \gamma (\log p^{+}(v| c_t){-}\log p^{-}(v| c_t)), \notag 
\end{align}
where we omit the constant normalization term. Since LLMs parameterize the next-token distribution $p_\gamma(\cdot | c_t) $ through their logits $z$ before the softmax layer, i.e., $p_\gamma(\cdot | c_t) = \text{softmax}(z_t^\top)$, we let $z_t^0$, $z_t^{+}$, and $z_t^{-}$ denote the unconditioned, positive-conditioned, and negative-conditioned logits at context $c_t$. We compute the steered logits as
\begin{align}
    z_t^T &= z_t^0+\gamma(z_t^{+} - z_t^{-}). \label{eq:2} 
\end{align}
Such a procedure is closely related to \emph{classifier-free guidance} for controlling LLM generation~\citep{sanchez2024stay} and to logit-space steering methods for hallucination mitigation~\citep{zhao2025mitigating}.
In \methodname{}, we instantiate this contrastive direction using teacher
logits conditioned on verified correct and incorrect on-policy rollouts.

%% file: latex/4.Methods.tex
\section{Method}
\subsection{Motivation}
\label{sec:motiv}

We start this section by introducing the potential sources of this pattern mismatch as demonstrated in \Cref{fig:logprob_ill} and \Cref{fig:steering-vector}. This further motivates the corresponding solution that we introduce later.

The first source of mismatch arises when OPSD uses a dataset-provided reasoning path as privileged guidance. Since the teacher signal is anchored to a single reasoning path, the student may overfit to this particular response and its associated solution pattern. This motivates us to \textbf{aggregate} guidance over multiple verified positive contexts, rather than relying on a single privileged trajectory. However, directly augmenting the OPSD loss in Eq.~\eqref{eq:opsd-loss} with multiple positive responses is not straightforward, since the divergence $D(\cdot,\cdot)$ is nonlinear and therefore does not necessarily commute with averaging over different teachers' responses.

Second, as illustrated in \Cref{fig:steering-vector}, asking the teacher model to reflect on a provided reasoning path may induce response patterns that are specific to the self-reflection process. As shown in \Cref{fig:logprob_ill}, such patterns may suppress the model's original reasoning behaviors, such as deliberative expressions like \textit{``wait''} or \textit{``hold on''}, which can be useful for reasoning. This motivates \textbf{contrastive} guidance: since reflection-induced patterns are largely shared across successful and failed responses, contrasting them helps suppress shared artifacts while preserving outcome-discriminative guidance.
We defer the detailed setup and additional results for this pattern mismatch to \Cref{app:logprob_analysis}.

\subsection{Outcome-Labeled Rollouts}
In \methodname{}, we use verifiable outcome feedback to construct outcome-contrastive teacher guidance for OPSD. For each input $x_i$, we sample $G$ rollouts from the current policy, denoted by $\{y_j\}_{j=1}^{G}
    \sim
    \pi_\theta(\cdot \mid x_i)$
and verify each rollout using the ground-truth label or a rule-based
verifier provided by the dataset. This gives an outcome-labeled set
\begin{equation}
    \mathcal{Y}(x_i)
    =
    \{(y_j,r_j)\}_{j=1}^{G}, r_j \in \{0, 1\}, \notag 
\end{equation}
where $r_j$ indicates whether $y_j$ is correct. We then
partition the rollouts into positive and negative sets:
\begin{equation}
\begin{aligned}
    \mathcal{Y}^{+}(x_i)
    &=
    \{y_j \mid r_j=1,\ j\in[G]\}, \\
    \mathcal{Y}^{-}(x_i)
    &=
    \{y_j \mid r_j=0,\ j\in[G]\}. \notag 
\end{aligned}
\end{equation}
Here, $\mathcal{Y}^{+}(x_i)$ contains verified correct rollouts,
while $\mathcal{Y}^{-}(x_i)$ contains incorrect rollouts. When clear
from context, we omit the dependence on $x_i$.

\subsection{Maintaining Mixed Guidance Logits}
The outcome-labeled rollouts above provide multiple contexts for querying the privileged teacher. Instead of using only the dataset-provided reference solution, we aggregate teacher guidance from verified correct and incorrect rollouts. This gives positive and negative teacher signals that can be contrasted in the steering step.

For an input $x_i$, the positive guidance pool consists of the correct rollouts $\mathcal Y^+$ and the dataset-provided reference solution $s_i$:
\begin{equation}
\begin{aligned}
    \mathcal{P}_{j}^{+}(x_i)
    =
    \mathcal{Y}^{+}(x_i) \cup \{s_i\}, \notag 
\end{aligned}
\end{equation}
Specifically, if no dataset-provided solution $s_i$ is available, we will use only verified correct rollouts as the positive guidance pool. Similarly, the negative guidance pool is defined as the set of incorrect rollouts, i.e., $\mathcal{P}_{j}^{-}(x_i)=\mathcal{Y}^{-}(x_i)$.

Following the composition of steering vector defined in Eq.~\eqref{eq:2}, the positive and negative guidance logits can therefore be computed as 
\begin{align}
\bar{z}^{+}_{j,t} &= \frac{1}{|\mathcal{P}_{j}^{+}|}
\textstyle{\sum_{g \in \mathcal{P}_{j}^{+}}}
z_{\theta'}(\cdot \mid x_i, g, y_{j,<t}). \notag \\
\bar{z}^{-}_{j,t} &= \frac{1}{|\mathcal{P}_{j}^{-}|}
\textstyle{\sum_{g \in \mathcal{P}_{j}^{-}}}
z_{\theta'}(\cdot \mid x_i, g, y_{j,<t}).
    \label{eq:get-mean-dist}
\end{align}

This logit-space aggregation can be viewed as a practical surrogate for averaging teacher guidance over multiple privileged contexts, while remaining
directly compatible with the logit-space steering operations. Intuitively, it constructs positive and negative averaged logits from correct and incorrect
contexts, reducing the effect of single-solution bias.

\subsection{Outcome-Guided Logit Steering}
Given the mixed guidance logits $\bar{z}^{+}_{j,t}$ and $\bar{z}^{-}_{j,t}$, constructed by averaging over multiple responses and reasoning paths generated by the student model as well as the dataset-provided solution, we define the \textbf{outcome-guided steered logits} as
\begin{align}
z^{T}_{j,t} = z_{\theta'}(\cdot \mid x_i, y_{j,<t})  +
\lambda (\bar{z}^{+}_{j,t}
-
\bar{z}^{-}_{j,t}),
\end{align}
where $\lambda \ge 0$ controls the strength of the steering signal. Here, the contrastive logit direction $\bar{z}^{+}_{j,t}-\bar{z}^{-}_{j,t}$ extracts the guidance signal by contrasting successful and failed reasoning paths. Since shared artifacts introduced by privileged prompting or self-reflection are expected to appear in both $\bar{z}^{+}_{j,t}$ and $\bar{z}^{-}_{j,t}$, taking their difference helps suppress these reflection-specific patterns. Adding this direction to the base logits $z_{\theta'}(\cdot \mid x_i, y_{j,<t})$ therefore calibrates the teacher signal while preserving the model's original on-policy response distribution. The steered teacher distribution is then derived by taking the softmax of the logits as \begin{align}
   \pi_T(\cdot \mid x_i, y_{j,<t}) = \mathrm{softmax}(z^{T}_{j,t}).\label {eq:teacher-dist} 
\end{align}

Similar to OPSD, we train the student by matching its token-level distribution $\pi_{\theta}(\cdot \mid x_i, y_{j,<t})$ to the steered teacher distribution $\pi_T(\cdot \mid x_i, y_{j,<t})$ along each student-generated response $y_j$ via
\begin{equation}
\begin{aligned}
&L_j^{\mathrm{steer}}(\theta)
= \frac{1}{H_j}
\\
&\;\cdot \sum_{t=1}^{H_j}\!D\big(
\mathrm{sg}[
\pi_T(\cdot | x_i, y_{j,<t})
],
\pi_{\theta}(\cdot | x_i, y_{j,<t}) 
\big),
\end{aligned}
\label{eq:seq-steer-loss}
\end{equation}
where $H_j=\min(|y_j|,H)$ is the effective distillation horizon, and $H$ is the maximum number of rollout tokens for this distribution-matching loss. 

Since the positive--negative contrast is intended to attenuate shared privileged-pattern shifts, we treat the resulting direction as a more outcome-discriminative teacher signal and use $\lambda>1$ to strengthen it. However, for rollouts that are already correct, additional full-vocabulary steering may unnecessarily reshape the model's token distribution and perturb its correct behavior. Therefore, we apply the steering loss only to incorrect rollouts:
\begin{equation}
\mathcal{L}_{\mathrm{steer}}(\theta)
=
\frac{1}{|\mathcal{Y}^{-}(x_i)|}
\sum_{y_j\in\mathcal{Y}^{-}(x_i)}
L_j^{\mathrm{steer}}(\theta),
\label{eq:steer-loss}
\end{equation}
with $\mathcal{L}_{\mathrm{steer}}(\theta)=0$ when $\mathcal{Y}^{-}(x_i)=\emptyset$.
\paragraph{Length-stabilizing positive regularization.}
In practice, applying steering only to incorrect rollouts can sometimes induce
length drift, causing generations to become increasingly long and exceed the
training-time rollout budget before producing a final answer. To stabilize training, we add a lightweight regularization term on short verified positive rollouts. Specifically, for positive rollouts whose length is below a fixed threshold $T_{\max}^{+}$, we apply a standard token-level SFT loss only on the
last $K$ tokens:
\begin{equation}
\begin{aligned}
\mathcal{L}_{\mathrm{pos}}(\theta)
&=
\frac{1}{|\mathcal{J}_+|}
\sum_{j\in\mathcal{J}_+}
\\
&\quad
\sum_{\mathclap{t=\max(1,|y_j|-K+1)}}^{|y_j|}
-\log \pi_\theta(y_{j,t}\mid x_i,y_{j,<t}),
\end{aligned}
\label{eq:pos-tail-loss}
\end{equation}
where $\mathcal{J}_{+}=\{j\in[G]: r_j=1,\ |y_j|\le T_{\max}^{+}\}$.
We use $K=128$ in our experiments. This regularizer does not provide full-vocabulary guidance; instead, it anchors the model on already-correct short trajectories, encouraging the model to produce final answers in the desired format while mitigating length drift. The final training objective for prompt $x_i$ is
\begin{equation}
\mathcal{L}(\theta)
=
\mathcal{L}_{\mathrm{steer}}(\theta)
+
\eta \mathcal{L}_{\mathrm{pos}}(\theta),
\label{eq:final-loss}
\end{equation}
where $\eta$ controls the strength of the positive regularization. We provide additional discussion of this regularizer in \Cref{app:pos_tail_eff} and summarize the overall procedure in \Cref{alg:method} in the appendix.

%% file: latex/5.Experiments.tex
\section{Experiments}

\subsection{Setup}

We evaluate \methodname{} on Qwen3-1.7B and Qwen3-4B~\citep{qwen3}, following
the original OPSD setting~\citep{opsd}. We train on OpenThought~\citep{openthought}
and evaluate on several challenging mathematical reasoning benchmarks, including
AIME 2024, AIME 2025, AMC23, HMMT, and MATH500~\citep{aime24,aime25,maa_amc,hmmt,hendrycks2021math}.
For \methodname{}, we sample 8 rollouts for each question, instantiate the distribution-matching loss \(D\) as forward KL, apply symmetric per-token pointwise divergence clipping, and use a ramp-up schedule for the steering
coefficient \(\lambda\), with a maximum value of 3. Following the common self-distillation evaluation setting, we set the maximum generation length to 38,912 tokens during evaluation. During training, we generate rollouts with a maximum length of 8,192 tokens to obtain verifier outcomes. Following the OPSD framework, we compute the distillation loss only on the first 1,024 tokens of each rollout. All methods are trained with LoRA using the same backbone model and training configuration for a fair comparison. We defer the detailed implementation settings, including distillation length, teacher update scheme, and positive SFT regularization to \Cref{app:exp_details}. \looseness=-1
\begin{table*}[t]
\centering
\small
\caption{
Main results of Qwen3-1.7B/4B on five reasoning benchmarks.
All scores are reported as Avg@8.
}
\vspace{-1em}
\label{tab:main_results}
\setlength{\tabcolsep}{6pt}
\renewcommand{\arraystretch}{1.12}
\begin{tabular}{lcccccc}
\toprule
\textbf{Method} & \textbf{AIME24} & \textbf{AIME25} & \textbf{AMC23} &
\textbf{HMMT} & \textbf{MATH500} & \textbf{Avg.} \\
\midrule
\multicolumn{7}{l}{\textit{Qwen3-1.7B}} \\
Base              & 51.5 & 36.7 & 86.3 & 24.6 & 91.2 & 58.1 \\
+ SFT             & 48.4 & 36.3 & 85.0 & 26.7 & 91.3 & 57.5 \\
+ GRPO            & 51.1 & 38.3 & 84.7 & 22.3 & 91.3 & 57.5 \\
+ OPSD            & 56.3 & 45.8 & 86.6 & 29.6 & 92.4 & 62.1 \\
+ Routed OPSD/GRPO & 57.9 & 38.7 & 85.6 & 25.0 & 92.3 & 59.9 \\
\rowcolor{gray!12}
+ \methodname{}   & \textbf{61.7} & \textbf{48.3} & \textbf{90.6} &
\textbf{30.0} & \textbf{93.0} & \textbf{64.8} \\
\midrule
\multicolumn{7}{l}{\textit{Qwen3-4B}} \\
Base              & 72.9 & 66.4 & 97.5 & 42.5 & 95.3 & 74.9 \\
+ SFT             & 70.2 & 62.3 & 93.1 & 39.2 & 93.0 & 71.6 \\
+ GRPO            & 75.6 & 68.1 & 97.5 & 44.2 & 95.3 & 76.1 \\
+ OPSD            & 76.3 & 69.1 & 96.3 & 44.2 & \textbf{95.5} & 76.3 \\
\rowcolor{gray!12}
+ \methodname{}   & \textbf{78.8} & \textbf{72.5} & \textbf{98.4} &
\textbf{46.7} & 95.3 & \textbf{78.3} \\
\bottomrule
\end{tabular}
\end{table*}
\subsection{Baselines}

We compare against three main baselines. \textbf{SFT} directly imitates expert trajectories from the dataset. \textbf{GRPO}~\citep{grpo} optimizes the policy using outcome-level rewards from verified rollouts. \textbf{OPSD}~\citep{opsd} performs on-policy self-distillation with privileged teacher guidance. For Qwen3-1.7B, we also include an outcome-routed hybrid baseline that applies the OPSD loss to failed rollouts and the GRPO loss to successful rollouts. This baseline provides a hybrid of RLVR and privileged distillation, inspired
by recent RL-distillation hybrid methods~\citep{ding2026hdpo,unify-grpo-sdpo}. Since exact reproduction requires method-specific engineering choices and
public implementations are currently unavailable, we evaluate this hybrid
baseline using our own implementation under the same training setup as the
other baselines.



\subsection{Main Results}

As shown in \Cref{tab:main_results}, \methodname{} consistently outperforms the baselines on most benchmarks, demonstrating the effectiveness of outcome-guided contrastive steering for reasoning tasks.

\begin{figure}[t]
    \centering
    \begin{subfigure}{0.48\linewidth}
        \centering
        \includegraphics[width=\linewidth]{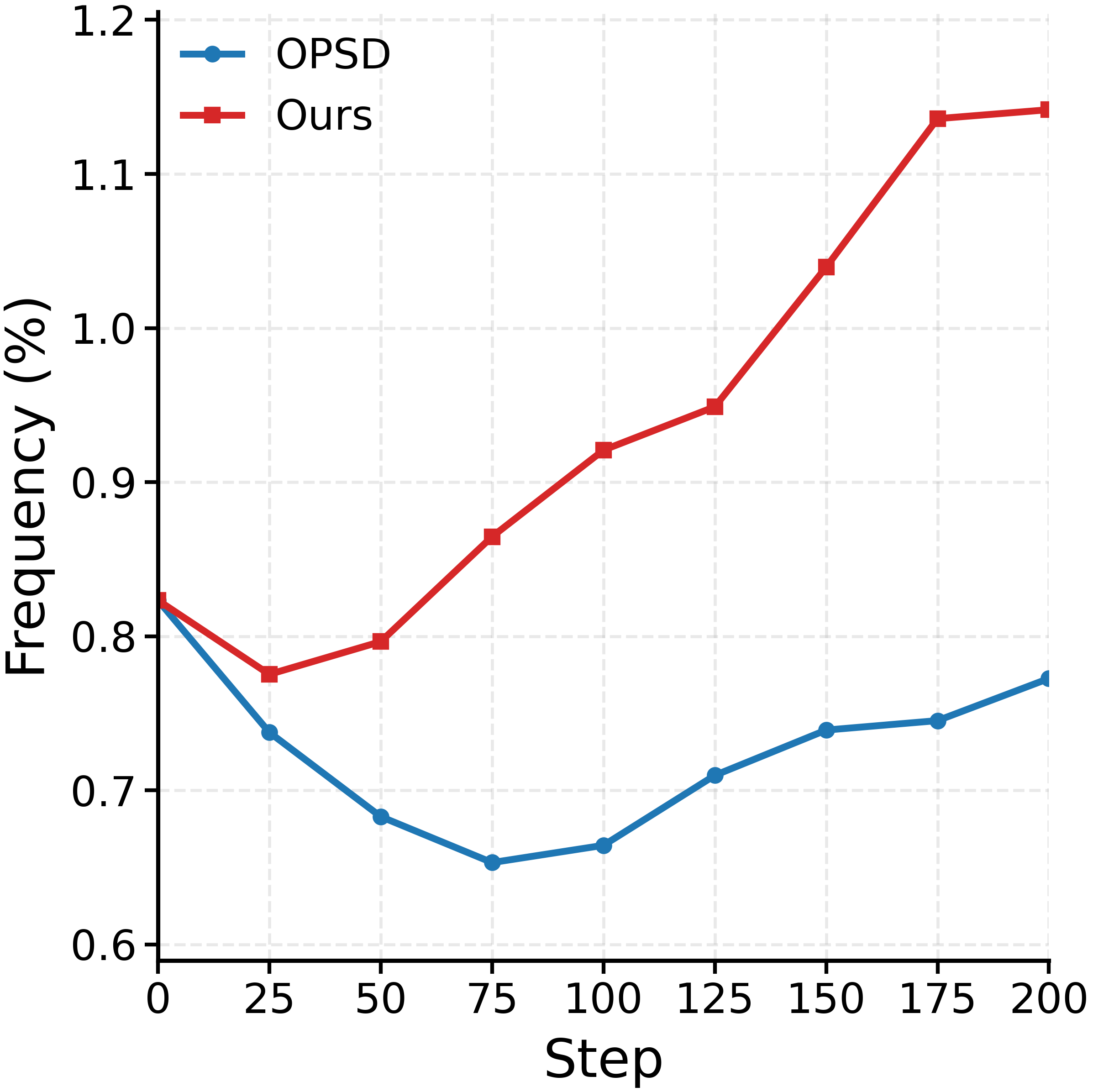}
        \caption{Qwen3-1.7B}
        \label{fig:wordcount-1p7b}
    \end{subfigure}
    \hfill
    \begin{subfigure}{0.48\linewidth}
        \centering
        \includegraphics[width=\linewidth]{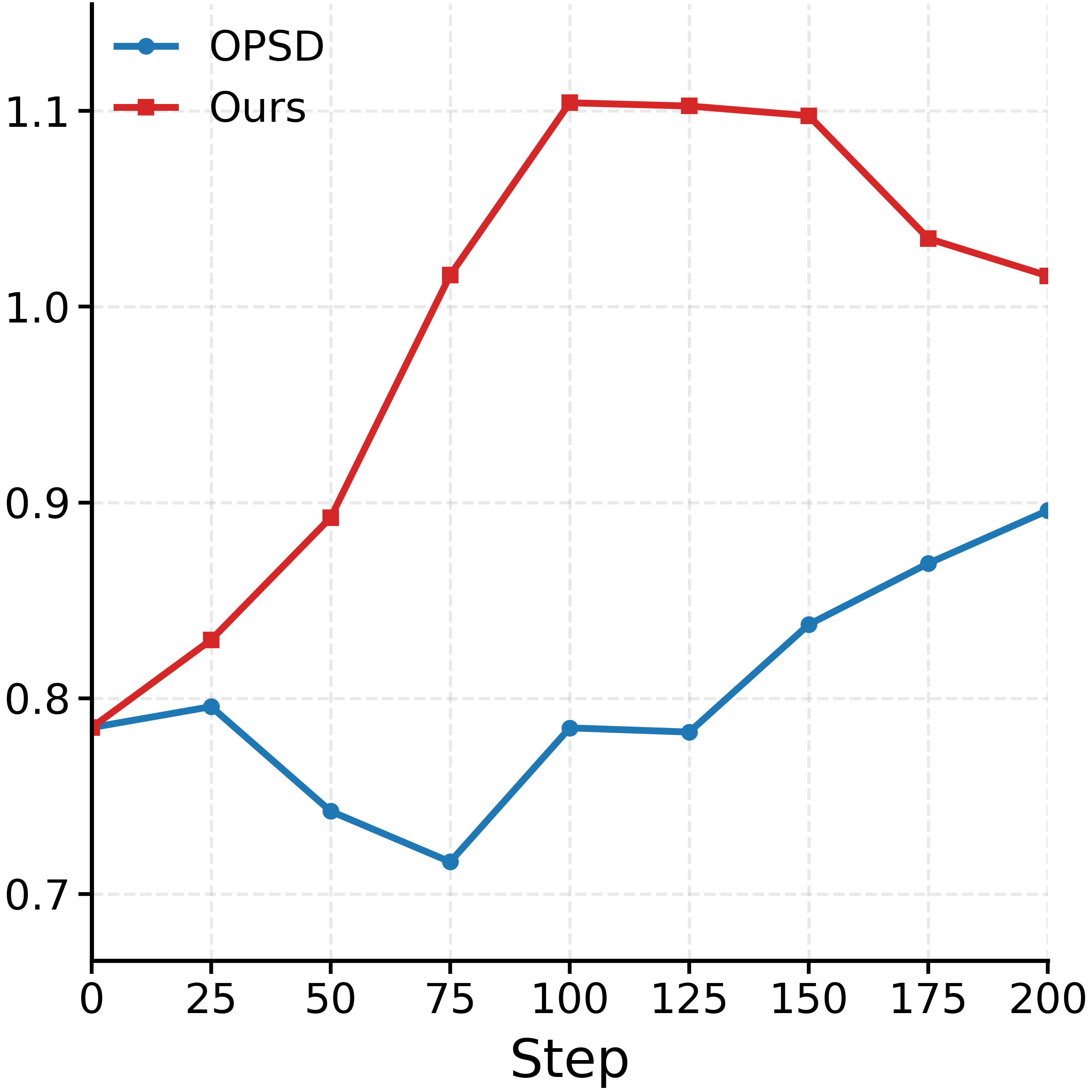}
        \caption{Qwen3-4B}
        \label{fig:wordcount-4b}
    \end{subfigure}
    \vspace{-.5em}
    \caption{
    Frequency of explicit epistemic reflection markers in reasoning traces on
    AIME 2024. Results are shown for (a) Qwen3-1.7B and (b) Qwen3-4B.
    }
    \label{fig:wordcount}
    \vspace{-.5em}
\end{figure}

To better understand how different objectives affect response-level reasoning patterns, we count the occurrence frequency of seven epistemic reflection markers in generated reasoning traces. As shown in \Cref{fig:wordcount}, OPSD initially reduces the frequency of these markers and only gradually recovers after around 100 training steps, but remains consistently lower than
\methodname{}. This suggests that \methodname{} better preserves markers associated with explicit self-checking and reconsideration. The full marker list is provided in \Cref{tab:reflection-keywords} in the appendix. We also provide a case analysis in \Cref{app:case_ana}.

\begin{figure}[t]
    \centering
    \includegraphics[width=\linewidth]{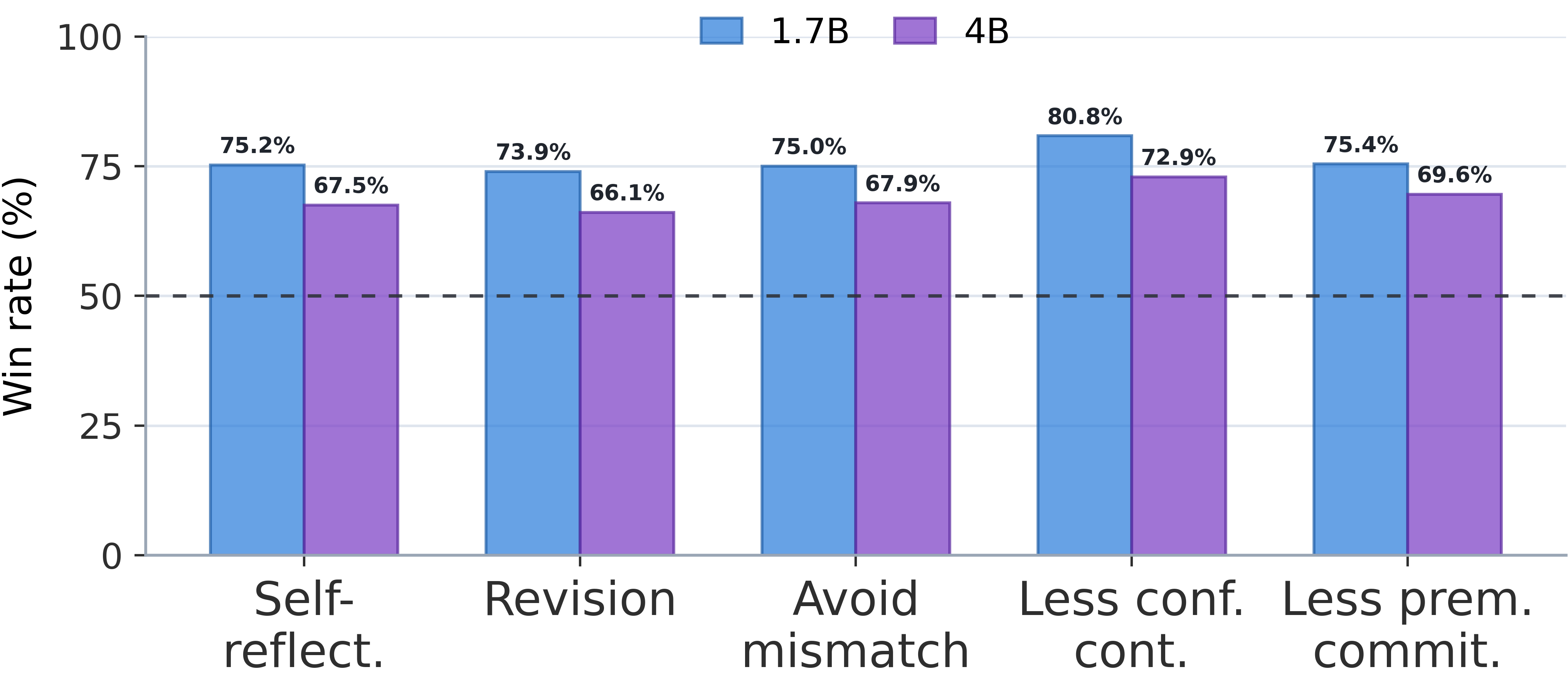}
    \vspace{-2em}
    \caption{
    Pairwise LLM-based behavioral annotation on AIME24@8. Each bar reports the percentage of pairwise comparisons in which \methodname{} is preferred over OPSD, after orienting all dimensions so that higher values indicate more desirable reasoning behavior. The dashed line marks the $50\%$ preference baseline. 
    }
    \label{fig:llm_judge}
\end{figure}

We further conduct a pairwise LLM-based behavioral annotation on AIME24.
For each problem, we use GPT-5.5~\citep{gpt5} as the judge to compare two anonymized
trajectories generated by OPSD and \methodname{} along five related behavioral dimensions: self-reflection, revision/backtracking, confident continuation,
premature commitment, and overconfident pattern mismatch. The annotation settings and full judge prompt are provided in \Cref{app:llm_judge,app:prompts}. As shown in \Cref{fig:llm_judge}, \methodname{} is preferred more often across the annotated behavioral dimensions. In particular, \methodname{} is more frequently judged to exhibit stronger self-reflection and more revision/backtracking, while being less associated with confident continuation and premature commitment. These results suggest that outcome-guided contrastive steering better preserves
self-checking and correction behavior. In contrast, OPSD-trained trajectories are more often associated with premature and overconfident reasoning patterns.

\begin{figure}[t]
\centering
\begin{subfigure}[t]{0.485\linewidth}
\centering
\includegraphics[width=\linewidth]{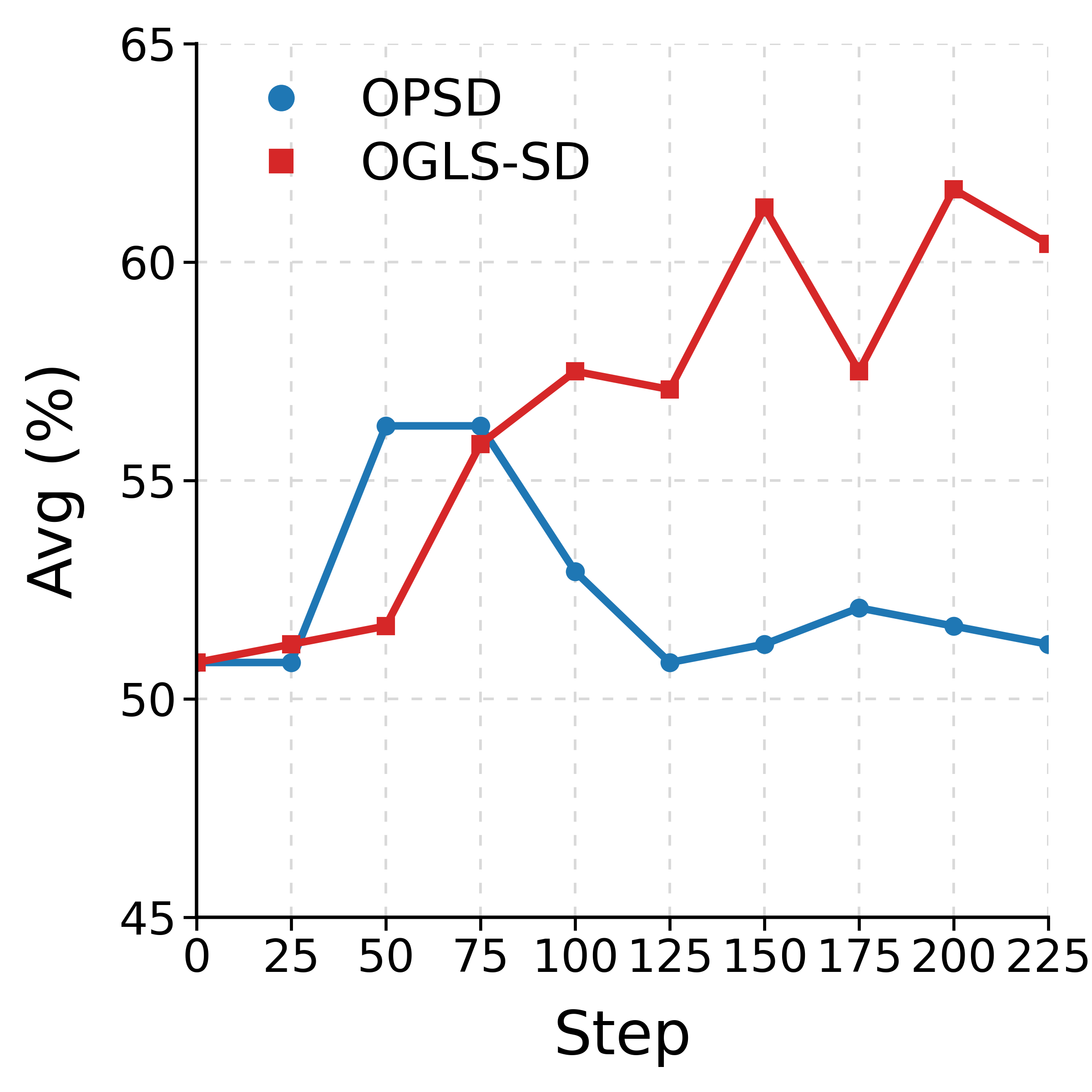}
\caption{Training dynamics.}
\label{fig:trainingcurve}
\end{subfigure}
\hfill
\begin{subfigure}[t]{0.485\linewidth}
\centering
\includegraphics[width=\linewidth]{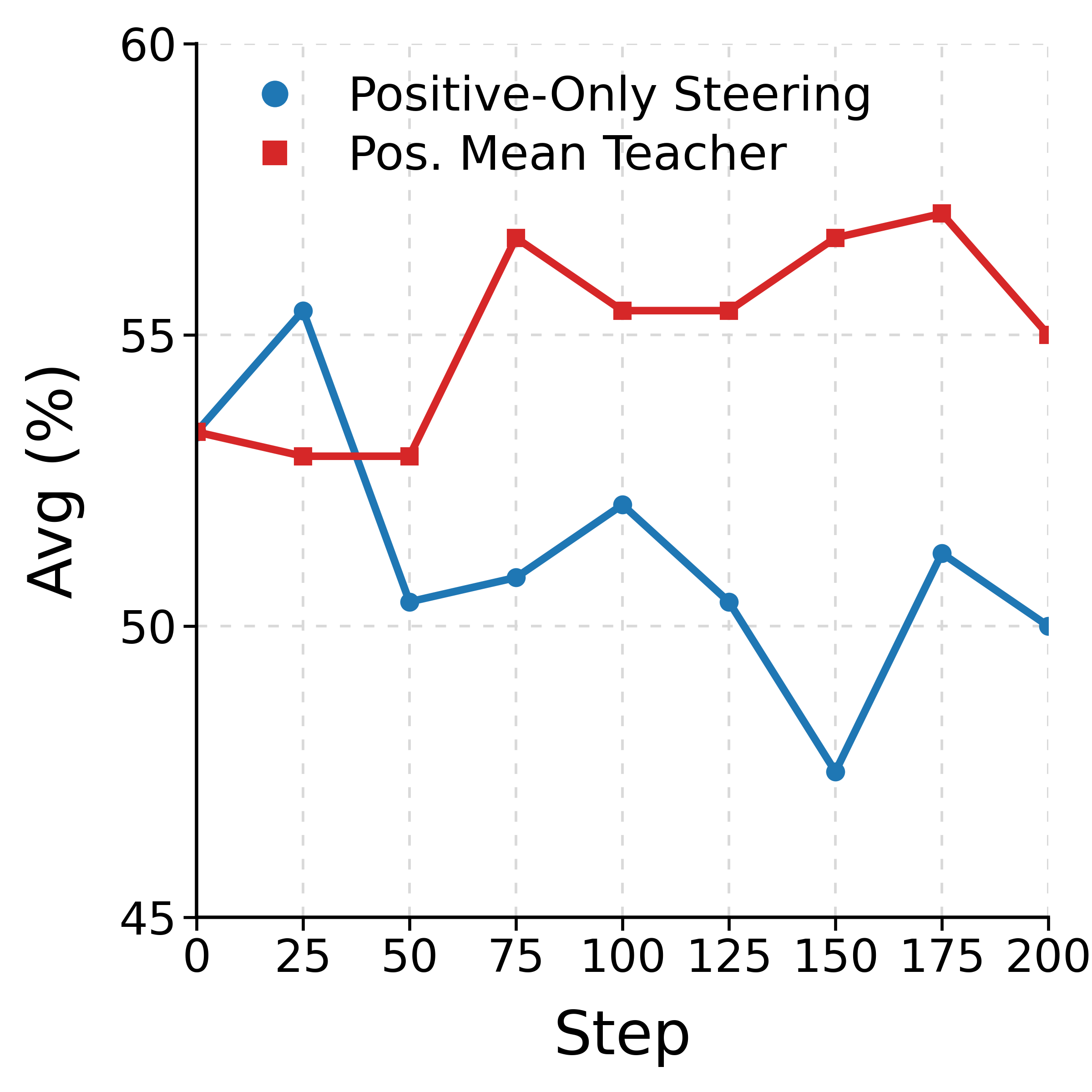}
\caption{Ablation study.}
\label{fig:ablation-curve}
\end{subfigure}
\vspace{-0.5em}
\caption{
Training dynamics and ablations on AIME24.
(a) OPSD peaks early and then degrades, while \methodname shows more stable improvement across training. (b) Ablations show that positive-mean guidance and positive-only steering
underperform \methodname{}.
}
\label{fig:curves}
\end{figure}

\subsection{Ablation Study}

To understand where the improvement of \methodname{} comes from, we conduct
ablation experiments on Qwen3-1.7B using AIME 2024, as shown in
\Cref{tab:ablation} and \Cref{fig:ablation-curve}. We focus on two questions: (1) Does the gain mainly come from using multiple positive reasoning paths instead of a single privileged solution? (2) Does the gain simply come from amplifying a positive steering direction with \(\lambda > 1\), rather than from explicitly contrasting positive and negative privileged guidance?

\vspace{.2em}\noindent \textbf{Does positive averaging suffice?}
To answer this question, we consider \textit{Pos. Mean Teacher}, which replaces the single privileged solution in OPSD with an averaged positive teacher distribution. Specifically, we apply an OPSD-like objective 
\begin{equation*}
\mathcal{L}_{\mathrm{PMT}}(\theta)
=
\tfrac{1}{H_j}\textstyle{
\sum_{t=1}^{H_j}}
D\big(\mathrm{sg}[\bar{\pi}_{\theta',t}^{+}],
    \pi_{\theta,t}
\big),
\end{equation*}

\noindent where
$\bar{\pi}_{\theta',t}^{+}=\bar{\pi}_{\theta'}^{+}(\cdot\mid x_i,y_{j,<t})$ is the averaged positive teacher distribution constructed from the positive guidance pool.
As shown in \Cref{tab:ablation}, this variant slightly improves over OPSD and
produces a relatively stable test curve in \Cref{fig:ablation-curve}, but it
still underperforms \methodname{}. This suggests that averaging multiple
positive guidance helps reduce solution-specific bias, but is
insufficient by itself.

\vspace{.2em}\noindent \textbf{Is positive-only steering enough?} We consider \textit{Positive-Only Steering}, which amplifies the positive guidance direction without using negative rollouts. The teacher logits are constructed as \vspace{-.3em}
\begin{equation*}
    z^{T}_{j,t}
    = z^{0}_{j,t} +\lambda (z^{+}_{j,t} - z^{0}_{j,t}),
\end{equation*}

\vspace{-.3em}\noindent where \(z^{0}_{j,t}\) denotes the base teacher logits and \(z^{+}_{j,t}\) denotes the positive teacher logits. This variant strengthens the positive guidance signal, but does not explicitly remove the shared privileged-pattern shift that may be present in positive-conditioned teacher guidance. As shown in \Cref{fig:ablation-curve}, Positive-Only Steering peaks early and then gradually degrades, indicating that simply amplifying positive guidance is insufficient for stable improvement.

We also include a variant that reduces the number of rollouts per question
from \(G=8\) to \(G=4\). As expected, performance decreases slightly, but the
variant still outperforms the other ablations. This suggests that
\methodname{} remains effective even with fewer sampled rollouts.

Overall, these ablations show that the gains of \methodname{} are not simply due to mean-teacher aggregation or a larger steering coefficient. Instead, explicitly contrasting positive and negative rollout-induced teacher patterns is important for improving both performance and stability.

\begin{table}[t]
\centering
\small
\caption{Ablation results on AIME 2024 using Qwen3-1.7B. Results are reported as Avg@8.}
\label{tab:ablation}
\vspace{-1em}
\begin{tabular}{lc}
\toprule
\textbf{Method} & \textbf{AIME 2024} \\
\midrule
\textit{Qwen3-1.7B}
Base & 51.5 \\
+ OPSD & 56.3 \\
+ Pos. Mean Teacher & 57.1 \\
+ Positive-Only Steering & 55.4 \\
+ \methodname{} w/ \(G=4\) & 60.8 \\
+ \methodname{} w/ \(G=8\) & \textbf{61.7} \\
\bottomrule
\end{tabular}
\vspace{-.2em}
\end{table}

%% file: latex/6.Conclusion.tex
\section{Conclusion}

In this paper, we propose \methodname, an on-policy self-distillation framework that constructs outcome-guided logit steering from successful and failed
rollouts. Instead of directly distilling from a single privileged teacher distribution, \methodname averages teacher logits over outcome-conditioned
rollout pools and contrasts them to reduce solution-specific bias and shared privileged-pattern shifts. Experiments on mathematical reasoning benchmarks
show that \methodname improves over standard OPSD and leads to more stable self-distillation.

\section*{Limitations}

Although \methodname{} shows consistent improvements under the OPSD evaluation setting, evaluating it on larger models, different model families, and broader reasoning tasks would provide a more comprehensive assessment.

We also mainly follow the original OPSD training and evaluation setup to ensure a controlled comparison. Therefore, we leave a more exhaustive study of design
choices, such as alternative divergence objectives, EMA or synchronized teacher updates, and different steering-coefficient schedules, to future work.

\textbf{Potential risks.} Our method improves post-training for mathematical reasoning models, and we do not target sensitive user-facing applications. However, stronger reasoning models may still be misused in domains where automated problem solving can cause harm, and the method inherits the risks of the underlying base model and training data.

%% file: latex/7.Appendix.tex
\Crefname{tcolorbox}{box}{boxes}

\section{Additional Experiments and Analyses}
\begin{algorithm*}[t]
\caption{\methodname: Outcome-Guided Logit Steering}
\label{alg:method}
\KwIn{Training dataset $\mathcal{D}$, policy $\pi_\theta$, teacher $\pi_{\theta'}$,
verifier $R$, rollout number $G$, steering strength $\lambda$, training steps $N$.}
\KwOut{Updated policy $\pi_\theta$.}

\For{$i=1,\dots,N$}{
    Sample $(x_i,s_i)$ from $\mathcal{D}$

    Sample rollouts $\{y_j\}_{j=1}^{G}\sim \pi_\theta(\cdot\mid x_i)$\;

    Verify each rollout by $r_j\gets R(x_i,y_j)$ for $j=1,\dots,G$\;

    \For{$j=1,\dots,G$}{
     \If{$r_j=1$}{
        \If{$|y_j| \le T_{\max}^{+}$}{
            Accumulate $L_{\mathrm{pos}}$ in \Cref{eq:pos-tail-loss}\tcp*{Positive-tail SFT}
        }
        \textbf{continue}\tcp*{Skip steering}
    
    }

        Set $\mathcal{P}_{j}^{+}\gets \{y_k:r_k=1,\ k\in[G]\}\cup\{s_i\}$ 
        and $\mathcal{P}_{j}^{-}\gets \{y_k:r_k=0,\ k\in[G]\}$\;

        \If{$\mathcal{P}_{j}^{+}=\emptyset$ \textbf{or} $\mathcal{P}_{j}^{-}=\emptyset$}{
            \textbf{continue}\tcp*{Contrastive direction is undefined}
        }

        Compute mixed guidance logits
        $\{\bar z^{+}_{j,t},\bar z^{-}_{j,t}\}_{t=1}^{H_j}$
        over the sequence using \Cref{eq:get-mean-dist}\;

        Construct the steered teacher distribution
        $\pi_T(\cdot\mid x_i,y_{j,<t})$ using \Cref{eq:teacher-dist}\;

        Compute rollout steering loss $L_j^{\text{steer}}$ using \Cref{eq:seq-steer-loss}\;
    }

    Gather rollout steering losses $\{L_j^{\text{steer}}\}$ and compute the final loss
    $\mathcal{L}(\theta)$ using \Cref{eq:final-loss}\;

    Update $\pi_\theta$ by minimizing $\mathcal{L}(\theta)$\;

    Maintain teacher $\pi_{\theta'}$ according to the chosen teacher scheme;
}
\end{algorithm*}
\subsection{Case Analysis}
\label{app:case_ana}
\begin{figure*}[t]
\centering
\includegraphics[width=\linewidth]{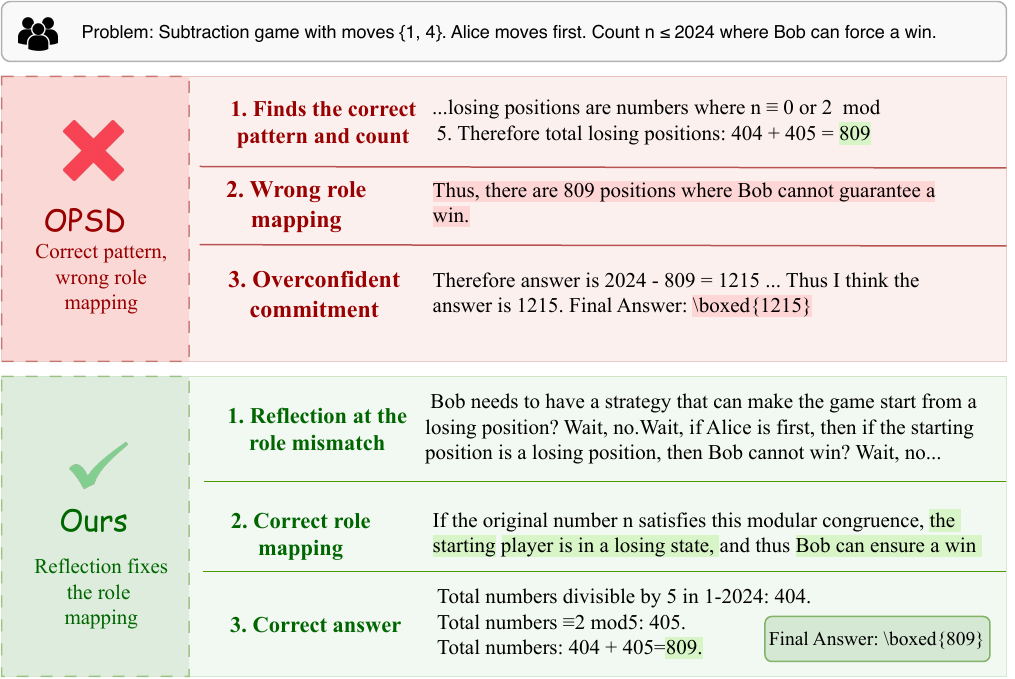}

\caption{
Case study comparing OPSD and \methodname{}. Both methods identify the
relevant losing-position structure, but OPSD commits to an incorrect
Alice/Bob role mapping without sufficient reflection. In contrast,
\methodname{} revisits the role assignment, checks the implication of the
losing positions, and preserves the correct answer.
}
\label{fig:casestudy}
\end{figure*}

We first provide a case analysis comparing OPSD and \methodname{}.
As shown in \Cref{fig:casestudy}, both methods identify the relevant losing-position structure, but both face a key Alice/Bob role-mapping issue. OPSD commits to an incorrect mapping without sufficient reflection, leading it
to draw the wrong conclusion. In contrast, \methodname{} explicitly revisits the role assignment, carefully checks the implication of the losing positions, and avoids the mistake.

\subsection{Diagnostic Analysis: Pattern Drift in Privileged Guidance}
\label{app:logprob_analysis}

\begin{table*}[hbt]
\centering
\caption{
Average token-level support change for common deliberative expressions on AIME24 using Qwen3-1.7B under two reference-model teacher constructions. Values are computed as
$\Delta_{\mathrm{supp}}
= \log \pi_T(y_t\mid x,y_{<t})
- \log \pi_{\mathrm{ref}}(y_t\mid x,y_{<t})$.
Negative values indicate reduced support relative to the unconditioned reference model.
}
\label{tab:reflect_logprob}
\begin{tabular}{lccc}
\toprule
Expression & Direct positive & Contrastive & Difference \\
\midrule
\textbf{hmm} & -1.4180 & +0.2264 & +1.6444 \\
\textbf{wait} & -0.7452 & +0.0429 & +0.7881 \\
\textbf{hold on} & -1.3078 & +0.3650 & +1.6728 \\
\textbf{let me check} & -0.3708 & +0.0276 & +0.3984 \\
\textbf{not quite} & +0.2203 & +0.0517 & -0.1686 \\
\textbf{let me rethink} & -0.1275 & +0.0390 & +0.1665 \\
\bottomrule
\end{tabular}
\end{table*}
We provide a reference-model diagnostic to illustrate how privileged guidance can alter token-level support for deliberative expressions. All teacher distributions are constructed from the same reference model, allowing us to isolate the effect of the teacher construction itself.

For each rollout token $y_t$, we compare the teacher log-probability with the
unconditioned reference log-probability:
\begin{align}
\Delta(y_t)
&=
\log \pi_T(y_t\mid x,y_{<t})
\nonumber\\
&\quad -
\log \pi_{\mathrm{ref}}(y_t\mid x,y_{<t}) .
\end{align}
Positive values indicate that the teacher increases the probability of the rollout token relative to the unconditioned reference model, while negative values indicate reduced support.

We compare two teacher constructions. The first directly conditions the
reference model on a correct guidance trace $G^+$:
\begin{equation}
\log \pi_T^{\mathrm{dir}}(\cdot \mid x,y_{<t})
=
\log \pi_{\mathrm{ref}}(\cdot \mid x,G^+,y_{<t}).
\end{equation}
The second uses a positive-negative contrastive construction in logit space:
\begin{align}
&\log \pi_T^{\mathrm{ctr}}(\cdot \mid x,y_{<t})
=
\log \pi_{\mathrm{ref}}(\cdot \mid  x,y_{<t})
\nonumber\\
&\quad
+ \lambda
\Big[
\log \pi_{\mathrm{ref}}(\cdot \mid x,G^+,y_{<t})
\nonumber\\
&\qquad\quad
-
\log \pi_{\mathrm{ref}}(\cdot \mid x,G^-,y_{<t})
\Big].
\end{align}
Here, $G^+$ and $G^-$ denote correct and incorrect guidance traces sampled
from the reference policy, respectively. We compute
$\Delta(y_t)$ on the same generated rollout tokens for both
teacher constructions.

As shown in \Cref{fig:logprob_ill}, directly conditioning on $G^+$ can reduce
support for deliberative tokens such as ``wait'' and ``hold on''. These tokens
often correspond to self-checking and revision behaviors in long-chain
reasoning. In contrast, the contrastive construction preserves or increases
support for these tokens by subtracting shared response-pattern shifts induced
by privileged conditioning. We further quantify this effect in
\Cref{tab:reflect_logprob}, where most deliberative expressions receive negative
support changes under direct positive guidance, but near-neutral or positive
support changes under contrastive guidance.

\paragraph{Diagnostic setup.}
We conduct this diagnostic on AIME 2024 using Qwen3-1.7B as the reference model. For each problem, we first sample 8 rollouts from the reference model with a
maximum completion length of 38k tokens, following the evaluation setting used in OPSD. We use the verifier to select one correct rollout and one incorrect rollout as the positive and negative guidance traces $G^+$ and $G^-$, respectively. We then evaluate the two teacher constructions on rollout tokens $y$ sampled separately from the guidance traces, using teacher forcing. This ensures that the direct positive-guidance and contrastive constructions are compared on the same token sequence. We set $\lambda=1$ for the contrastive
construction. Since Qwen3-1.7B supports a maximum context length of 40K tokens, the concatenation of the problem, an auxiliary guidance trace $G$, and the
local rollout prefix $y_{<t}$ may exceed the context window. In such cases, we truncate the middle part of the auxiliary guidance trace $G$, while preserving the problem statement, the final-answer region of $G$, and the local rollout prefix used for computing token probabilities.

\subsection{Effect of Positive-Tail Regularization}
\label{app:pos_tail_eff}

We further examine the role of the positive-tail SFT regularizer and show that it mainly stabilizes generation length rather than serving as the primary
learning signal. We ablate this lightweight regularizer in \methodname{} using Qwen3-1.7B on AIME24. We report both Avg@8 and the answer-format rate, defined as the percentage of generated responses that produce a final answer within the generation budget.
\begin{figure*}[htb]
    \centering
    \includegraphics[width=1\linewidth]{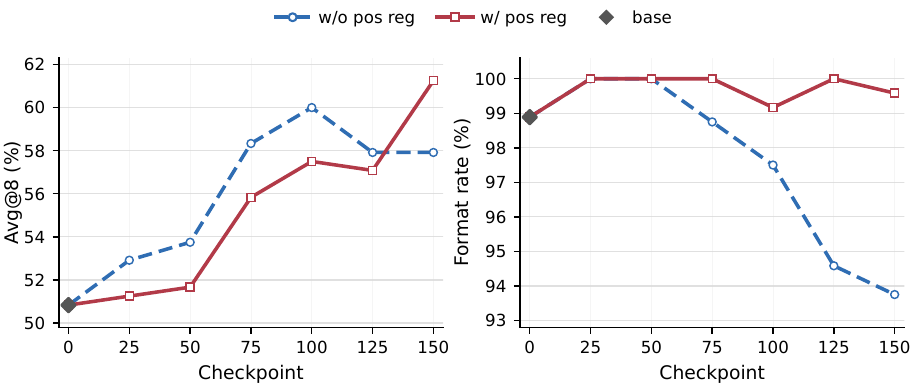}
    \caption{
    Effect of positive-tail regularization on AIME24 using Qwen3-1.7B. Removing the positive-tail regularizer yields even faster early gains, suggesting that the primary learning signal comes from outcome-guided  distillation. However, it also causes a steady drop in answer-format rate, indicating length drift and reduced final-answer completion within the generation budget. Positive-tail regularization stabilizes the answer-format rate and improves later-stage training stability.
    }
    \label{fig:pos-reg-analysis}
\end{figure*}
As shown in \Cref{fig:pos-reg-analysis}, removing positive-tail regularization leads to even faster performance gains in the early stage, despite a steadily decreasing answer-format rate. This suggests that the primary learning signal comes from the outcome-guided distillation objective rather than from the positive SFT term. In other words, the positive-tail SFT
term mainly acts as a stabilizing regularizer, trading off some early-stage aggressiveness for improved later-stage stability.

However, as training proceeds, the answer-format rate without positive-tail regularization continues to drop, indicating length drift and reduced final-answer completion within the generation budget. This eventually makes training less stable and hurts later checkpoints. In contrast, the positive-tail regularizer maintains a high answer-format rate throughout training and yields more stable final performance. These results support our design choice of using positive-tail SFT as a stabilization term, rather than as the primary source of performance gains.  Detailed hyperparameters for the positive-tail SFT regularizer are provided in \Cref{app:exp_details}.

\section{Experimental Details}
\label{app:exp_details}
\subsection{Implementation Details}

We first describe the implementation details of \methodname. We generally follow the official OPSD codebase~\citep{opsd} and build our method on top of it. We use LoRA~\citep{LoRA} for post-training, with learning rate $5\times 10^{-6}$, LoRA rank $r=64$, and LoRA scaling factor $\alpha=128$.

The key training and evaluation hyperparameters are summarized in
\Cref{tab:hyperparams}. We use vLLM~\citep{vllm} for rollout sampling and TRL~\citep{trl} for training. We set the training rollout budget to 8,192 tokens to ensure that most rollouts contain a final answer. Following OPSD, we set $H=1024$ and compute the distillation loss on at most the first 1,024 tokens of each trajectory, with $H_j=\min(|y_j|,H)$. We keep the teacher fixed as the original base model. For the positive SFT regularizer, we select verified correct rollouts whose length is below 4,096 tokens and apply SFT only to their last 128 tokens. This lightweight regularization is not intended to provide full-trajectory supervision; instead, it anchors the model on concise correct completions and mitigates length drift, where generations become increasingly long during training.
We use an adaptive coefficient for this auxiliary term:
$
\eta = \eta_0 \left(1 - \mathrm{Acc}_i\right),
$
where $\eta_0$ is a fixed hyperparameter and $\mathrm{Acc}=\frac{1}{G}\sum_{j=1}^{G} r_{ij}$ denotes the fraction of verified-correct on-policy rollouts among the $G$ samples for prompt $x_i$. This weighting gives larger relative weight to the selected positive-tail regularization when correct rollouts are scarce, while reducing its influence when the current policy already solves the prompt reliably. If no verified correct rollout is available, the positive regularization term is zero. We use a linear ramp-up schedule for the steering coefficient $\lambda$,
increasing it from 1 to 3 for Qwen3-1.7B and from 1.5 to 3 for Qwen3-4B.
We also apply symmetric pointwise divergence clipping with threshold 0.05.
During evaluation, the maximum generation budget is set to 38,912 tokens.
\paragraph{Computational cost.}
\methodname introduces additional training-time overhead compared with vanilla OPSD, because it requires teacher logit polling over positive and negative guidance pools. This overhead is incurred only during post-training and does not change the inference cost of the resulting model. In our experiments, long-context evaluation is also a major computational cost, since each model is evaluated with a maximum generation budget of 38k tokens. We therefore view OGLS-SD
as trading additional training-time computation for improved stability and reasoning performance.
\begin{table*}[hbt]
\centering
\caption{Key training and evaluation hyperparameters.}
\label{tab:hyperparams}
\begin{tabular}{lc}
\toprule
\textbf{Hyperparameter} & \textbf{Value} \\
\midrule
Rollouts per prompt $G$ & 8 \\
Sampling temperature & 1 \\
Training generation budget & 8,192 \\
Evaluation generation budget & 38,912 \\
Distillation length & first 1,024 tokens \\
Distribution-matching divergence $D$ & forward KL \\
Maximum steering coefficient $\lambda$ & 3 \\
Pointwise divergence clipping threshold & 0.05 \\
Positive rollout length threshold $T_{\max}^{+}$ & 4,096 \\
Positive-tail SFT length $K$ & 128 \\
Positive-tail SFT coefficient $\eta_0$ & 0.05 \\
Teacher model & fixed base model \\
Hardware & 4 NVIDIA A100 80GB GPUs \\
\bottomrule
\end{tabular}
\end{table*}

\subsection{LLM Judge Details}
\label{app:llm_judge}

We provide the implementation details for the LLM-based behavioral annotation
in \Cref{fig:llm_judge}. We compare trajectories generated by Qwen3 models
trained with OPSD and \methodname{} on AIME24.For each problem, we sample 8 rollouts from each model, form anonymized pairwise comparisons between OPSD and \methodname{} trajectories, and use the Azure OpenAI API deployment of \texttt{gpt-5.5-2026-04-24} to judge which trajectory better exhibits each behavioral dimension.

The judge compares five dimensions: stronger self-reflection, more revision/backtracking, less overconfident pattern mismatch, more confident continuation, and more premature commitment. The prompt intentionally includes both desirable and undesirable behavioral dimensions to discourage the judge from assigning all dimensions to the same trajectory by default. For stronger self-reflection, more revision/backtracking, and less overconfident pattern mismatch, we count the method selected by the judge as preferred. For more confident continuation and more premature commitment, which correspond to undesirable behaviors, we report the reverse preference so that the value reflects how often a method is judged to avoid the corresponding behavior.
Thus, larger values in \Cref{fig:llm_judge} consistently indicate more desirable reasoning behavior.

To align the annotation setting with the training-time rollout setting, we
disable thinking mode and use an 8,192-token generation budget. The full judge
prompt is provided in \Cref{fig:prompt_judge}.

\subsection{Prompt Details}
\label{app:prompts}
We first present the prompt format used for the privileged teacher in \Cref{fig:teacher-prompt}.
\input{latex/prompt/teacherprompt}

We also list the keywords used to measure explicit self-reflection frequency
in \Cref{tab:reflection-keywords}.

\begin{table}[t]
\centering
\caption{Keyword set for measuring explicit self-reflection frequency.}
\label{tab:reflection-keywords}
\begin{tabular}{l}
\toprule
\textbf{Keywords} \\
\midrule
\texttt{wait}, \texttt{hold on}, \texttt{oops}, \texttt{not quite}, \texttt{hmm} \\
\texttt{let me rethink}, \texttt{let me reconsider}\\ \texttt{let me check} \\
\bottomrule
\end{tabular}
\end{table}

The prompt used for pairwise LLM-based behavioral annotation is shown \Cref{fig:prompt_judge}. The judge compares two anonymized rollouts along five behavioral dimensions: stronger self-reflection, more
confident continuation, more premature commitment, more revision or
backtracking, and less overconfident pattern mismatch. For interpretation,
stronger self-reflection, more revision/backtracking, and less overconfident
pattern mismatch are preferred, while more confident continuation and more
premature commitment indicate less desirable response patterns.

\input{latex/prompt/judgeprompt}

%% file: latex/prompt/teacherprompt.tex
\begin{figure*}[hbt]
\centering
\begin{tcblisting}{
    title={Prompts for teacher model},
    colback=gray!5,
    colframe=gray!50,
    boxrule=0.5pt,
    listing only,
    listing options={basicstyle=\small\ttfamily,breaklines=true}
}
Problem: {problem}

Here is a reference solution to this problem:

=== Reference Solution Begin ===

{solution}

=== Reference Solution End ===

After reading the reference solution above, make sure you truly understand the reasoning behind each step--do not copy or paraphrase it. Now, using your own words and independent reasoning, derive the same final answer to the problem above. Think step by step, explore different approaches, and do not be afraid to backtrack or reconsider if something does not work out:

Please reason step by step, and put your final answer within \boxed{}.

\end{tcblisting}
\caption{Prompt used for the teacher model.}
\label{fig:teacher-prompt}
\end{figure*}

%% file: latex/prompt/judgeprompt.tex
\begin{figure*}
\centering
\begin{tcblisting}{
    title={Prompt for LLM Judge},
    colback=gray!5,
    colframe=gray!50,
    label={box:judge-prompt},
    boxrule=0.5pt,
    listing only,
    listing options={basicstyle=\small\ttfamily,breaklines=true}
}
You are given one math problem and two model-generated reasoning trajectories for the same problem.
Your task is NOT to solve the problem and NOT to judge whether the final answers are correct.
Your task is to compare the reasoning behavior shown in the two trajectories.
Definitions:
Self-reflection means the trajectory explicitly examines its own reasoning process, checks a previous step, questions an assumption, revises an intermediate conclusion, or backtracks after noticing a possible issue.
Confident continuation means the trajectory keeps advancing conclusions in a certain tone without meaningful checking, reconsideration, or uncertainty, especially when making nontrivial transitions.
Overconfident pattern mismatch means the trajectory appears to commit to a reasoning pattern, role/condition mapping, algebraic setup, or intermediate conclusion with high confidence before doing enough self-checking, and does not adequately revise or verify that commitment.
Compare Trajectory A and Trajectory B along the following dimensions:

1. stronger_self_reflection:
Which trajectory shows stronger self-reflection, self-checking, reconsideration, or correction?
Answer "A", "B", or "tie".

2. more_confident_continuation:
Which trajectory shows more confident, uninterrupted continuation without meaningful self-checking?
Answer "A", "B", or "tie".

3. more_premature_commitment:
Which trajectory shows more premature commitment to intermediate/final conclusions before checking assumptions, edge cases, or role/condition mappings?
Answer "A", "B", or "tie".

4. more_revision_or_backtracking:
Which trajectory shows more explicit revision, correction, backtracking, or validation after reconsidering?
Answer "A", "B", or "tie".

5. less_overconfident_pattern_mismatch:
Which trajectory better avoids overconfident pattern mismatch?
Answer "A", "B", or "tie".

Return only valid JSON with the following keys:
{
  "stronger_self_reflection": "A/B/tie",
  "more_confident_continuation": "A/B/tie",
  "more_premature_commitment": "A/B/tie",
  "more_revision_or_backtracking": "A/B/tie",
  "less_overconfident_pattern_mismatch": "A/B/tie",
  "brief_reason": "..."
}

Math problem:
[PROBLEM]
Reference answer:
[ANSWER]
Trajectory A:
[TRAJECTORY_A]
Trajectory B:
[TRAJECTORY_B]
\end{tcblisting}
\caption{Prompt Used for the LLM Judge.}
\label{fig:prompt_judge}
\end{figure*}

%% file: custom.bib
@misc{rethinkopd,
      title={Rethinking On-Policy Distillation of Large Language Models: Phenomenology, Mechanism, and Recipe}, 
      author={Yaxuan Li and Yuxin Zuo and Bingxiang He and Jinqian Zhang and Chaojun Xiao and Cheng Qian and Tianyu Yu and Huan-ang Gao and Wenkai Yang and Zhiyuan Liu and Ning Ding},
      year={2026},
      eprint={2604.13016},
      archivePrefix={arXiv},
      primaryClass={cs.LG},
      url={https://arxiv.org/abs/2604.13016}, 
}

@misc{rlsd,
      title={Self-Distilled RLVR}, 
      author={Chenxu Yang and Chuanyu Qin and Qingyi Si and Minghui Chen and Naibin Gu and Dingyu Yao and Zheng Lin and Weiping Wang and Jiaqi Wang and Nan Duan},
      year={2026},
      eprint={2604.03128},
      archivePrefix={arXiv},
      primaryClass={cs.LG},
      url={https://arxiv.org/abs/2604.03128}, 
}

@misc{opsd,
      title={Self-Distilled Reasoner: On-Policy Self-Distillation for Large Language Models}, 
      author={Siyan Zhao and Zhihui Xie and Mengchen Liu and Jing Huang and Guan Pang and Feiyu Chen and Aditya Grover},
      year={2026},
      eprint={2601.18734},
      archivePrefix={arXiv},
      primaryClass={cs.LG},
      url={https://arxiv.org/abs/2601.18734}, 
}

@misc{qwen3,
      title={Qwen3 Technical Report}, 
      author={An Yang and Anfeng Li and Baosong Yang and Beichen Zhang and Binyuan Hui and Bo Zheng and Bowen Yu and Chang Gao and Chengen Huang and Chenxu Lv and Chujie Zheng and Dayiheng Liu and Fan Zhou and Fei Huang and Feng Hu and Hao Ge and Haoran Wei and Huan Lin and Jialong Tang and Jian Yang and Jianhong Tu and Jianwei Zhang and Jianxin Yang and Jiaxi Yang and Jing Zhou and Jingren Zhou and Junyang Lin and Kai Dang and Keqin Bao and Kexin Yang and Le Yu and Lianghao Deng and Mei Li and Mingfeng Xue and Mingze Li and Pei Zhang and Peng Wang and Qin Zhu and Rui Men and Ruize Gao and Shixuan Liu and Shuang Luo and Tianhao Li and Tianyi Tang and Wenbiao Yin and Xingzhang Ren and Xinyu Wang and Xinyu Zhang and Xuancheng Ren and Yang Fan and Yang Su and Yichang Zhang and Yinger Zhang and Yu Wan and Yuqiong Liu and Zekun Wang and Zeyu Cui and Zhenru Zhang and Zhipeng Zhou and Zihan Qiu},
      year={2025},
      eprint={2505.09388},
      archivePrefix={arXiv},
      primaryClass={cs.CL},
      url={https://arxiv.org/abs/2505.09388}, 
}

@misc{mimo,
      title={MiMo-V2-Flash Technical Report}, 
      author={Core Team and Bangjun Xiao and Bingquan Xia and Bo Yang and Bofei Gao and Bowen Shen and Chen Zhang and Chenhong He and Chiheng Lou and Fuli Luo and Gang Wang and Gang Xie and Hailin Zhang and Hanglong Lv and Hanyu Li and Heyu Chen and Hongshen Xu and Houbin Zhang and Huaqiu Liu and Jiangshan Duo and Jianyu Wei and Jiebao Xiao and Jinhao Dong and Jun Shi and Junhao Hu and Kainan Bao and Kang Zhou and Lei Li and Liang Zhao and Linghao Zhang and Peidian Li and Qianli Chen and Shaohui Liu and Shihua Yu and Shijie Cao and Shimao Chen and Shouqiu Yu and Shuo Liu and Tianling Zhou and Weijiang Su and Weikun Wang and Wenhan Ma and Xiangwei Deng and Bohan Mao and Bowen Ye and Can Cai and Chenghua Wang and Chengxuan Zhu and Chong Ma and Chun Chen and Chunan Li and Dawei Zhu and Deshan Xiao and Dong Zhang and Duo Zhang and Fangyue Liu and Feiyu Yang and Fengyuan Shi and Guoan Wang and Hao Tian and Hao Wu and Heng Qu and Hongfei Yi and Hongxu An and Hongyi Guan and Xing Zhang and Yifan Song and Yihan Yan and Yihao Zhao and Yingchun Lai and Yizhao Gao and Yu Cheng and Yuanyuan Tian and Yudong Wang and Zhen Tang and Zhengju Tang and Zhengtao Wen and Zhichao Song and Zhixian Zheng and Zihan Jiang and Jian Wen and Jiarui Sun and Jiawei Li and Jinlong Xue and Jun Xia and Kai Fang and Menghang Zhu and Nuo Chen and Qian Tu and Qihao Zhang and Qiying Wang and Rang Li and Rui Ma and Shaolei Zhang and Shengfan Wang and Shicheng Li and Shuhao Gu and Shuhuai Ren and Sirui Deng and Tao Guo and Tianyang Lu and Weiji Zhuang and Weikang Zhang and Weimin Xiong and Wenshan Huang and Wenyu Yang and Xin Zhang and Xing Yong and Xu Wang and Xueyang Xie and Yilin Jiang and Yixin Yang and Yongzhe He and Yu Tu and Yuanliang Dong and Yuchen Liu and Yue Ma and Yue Yu and Yuxing Xiang and Zhaojun Huang and Zhenru Lin and Zhipeng Xu and Zhiyang Chen and Zhonghua Deng and Zihan Zhang and Zihao Yue},
      year={2026},
      eprint={2601.02780},
      archivePrefix={arXiv},
      primaryClass={cs.CL},
      url={https://arxiv.org/abs/2601.02780}, 
}

@misc{deepseekv4,
      title={DeepSeek-V4: Towards Highly Efficient Million-Token Context Intelligence},
      author={DeepSeek-AI},
      year={2026},
}

@misc{glm5,
      title={GLM-5: from Vibe Coding to Agentic Engineering}, 
      author={GLM-5-Team and : and Aohan Zeng and Xin Lv and Zhenyu Hou and Zhengxiao Du and Qinkai Zheng and Bin Chen and Da Yin and Chendi Ge and Chenghua Huang and Chengxing Xie and Chenzheng Zhu and Congfeng Yin and Cunxiang Wang and Gengzheng Pan and Hao Zeng and Haoke Zhang and Haoran Wang and Huilong Chen and Jiajie Zhang and Jian Jiao and Jiaqi Guo and Jingsen Wang and Jingzhao Du and Jinzhu Wu and Kedong Wang and Lei Li and Lin Fan and Lucen Zhong and Mingdao Liu and Mingming Zhao and Pengfan Du and Qian Dong and Rui Lu and Shuang-Li and Shulin Cao and Song Liu and Ting Jiang and Xiaodong Chen and Xiaohan Zhang and Xuancheng Huang and Xuezhen Dong and Yabo Xu and Yao Wei and Yifan An and Yilin Niu and Yitong Zhu and Yuanhao Wen and Yukuo Cen and Yushi Bai and Zhongpei Qiao and Zihan Wang and Zikang Wang and Zilin Zhu and Ziqiang Liu and Zixuan Li and Bojie Wang and Bosi Wen and Can Huang and Changpeng Cai and Chao Yu and Chen Li and Chengwei Hu and Chenhui Zhang and Dan Zhang and Daoyan Lin and Dayong Yang and Di Wang and Ding Ai and Erle Zhu and Fangzhou Yi and Feiyu Chen and Guohong Wen and Hailong Sun and Haisha Zhao and Haiyi Hu and Hanchen Zhang and Hanrui Liu and Hanyu Zhang and Hao Peng and Hao Tai and Haobo Zhang and He Liu and Hongwei Wang and Hongxi Yan and Hongyu Ge and Huan Liu and Huanpeng Chu and Jia'ni Zhao and Jiachen Wang and Jiajing Zhao and Jiamin Ren and Jiapeng Wang and Jiaxin Zhang and Jiayi Gui and Jiayue Zhao and Jijie Li and Jing An and Jing Li and Jingwei Yuan and Jinhua Du and Jinxin Liu and Junkai Zhi and Junwen Duan and Kaiyue Zhou and Kangjian Wei and Ke Wang and Keyun Luo and Laiqiang Zhang and Leigang Sha and Liang Xu and Lindong Wu and Lintao Ding and Lu Chen and Minghao Li and Nianyi Lin and Pan Ta and Qiang Zou and Rongjun Song and Ruiqi Yang and Shangqing Tu and Shangtong Yang and Shaoxiang Wu and Shengyan Zhang and Shijie Li and Shuang Li and Shuyi Fan and Wei Qin and Wei Tian and Weining Zhang and Wenbo Yu and Wenjie Liang and Xiang Kuang and Xiangmeng Cheng and Xiangyang Li and Xiaoquan Yan and Xiaowei Hu and Xiaoying Ling and Xing Fan and Xingye Xia and Xinyuan Zhang and Xinze Zhang and Xirui Pan and Xu Zou and Xunkai Zhang and Yadi Liu and Yandong Wu and Yanfu Li and Yidong Wang and Yifan Zhu and Yijun Tan and Yilin Zhou and Yiming Pan and Ying Zhang and Yinpei Su and Yipeng Geng and Yong Yan and Yonglin Tan and Yuean Bi and Yuhan Shen and Yuhao Yang and Yujiang Li and Yunan Liu and Yunqing Wang and Yuntao Li and Yurong Wu and Yutao Zhang and Yuxi Duan and Yuxuan Zhang and Zezhen Liu and Zhengtao Jiang and Zhenhe Yan and Zheyu Zhang and Zhixiang Wei and Zhuo Chen and Zhuoer Feng and Zijun Yao and Ziwei Chai and Ziyuan Wang and Zuzhou Zhang and Bin Xu and Minlie Huang and Hongning Wang and Juanzi Li and Yuxiao Dong and Jie Tang},
      year={2026},
      eprint={2602.15763},
      archivePrefix={arXiv},
      primaryClass={cs.LG},
      url={https://arxiv.org/abs/2602.15763}, 
}

@misc{sdpo,
      title={Reinforcement Learning via Self-Distillation}, 
      author={Jonas Hübotter and Frederike Lübeck and Lejs Behric and Anton Baumann and Marco Bagatella and Daniel Marta and Ido Hakimi and Idan Shenfeld and Thomas Kleine Buening and Carlos Guestrin and Andreas Krause},
      year={2026},
      eprint={2601.20802},
      archivePrefix={arXiv},
      primaryClass={cs.LG},
      url={https://arxiv.org/abs/2601.20802}, 
}

@misc{exopd,
      title={Learning beyond Teacher: Generalized On-Policy Distillation with Reward Extrapolation}, 
      author={Wenkai Yang and Weijie Liu and Ruobing Xie and Kai Yang and Saiyong Yang and Yankai Lin},
      year={2026},
      eprint={2602.12125},
      archivePrefix={arXiv},
      primaryClass={cs.LG},
      url={https://arxiv.org/abs/2602.12125}, 
}

@misc{unify-grpo-sdpo,
      title={Unifying Group-Relative and Self-Distillation Policy Optimization via Sample Routing}, 
      author={Gengsheng Li and Tianyu Yang and Junfeng Fang and Mingyang Song and Mao Zheng and Haiyun Guo and Dan Zhang and Jinqiao Wang and Tat-Seng Chua},
      year={2026},
      eprint={2604.02288},
      archivePrefix={arXiv},
      primaryClass={cs.LG},
      url={https://arxiv.org/abs/2604.02288}, 
}

@misc{grpo,
      title={DeepSeekMath: Pushing the Limits of Mathematical Reasoning in Open Language Models}, 
      author={Zhihong Shao and Peiyi Wang and Qihao Zhu and Runxin Xu and Junxiao Song and Xiao Bi and Haowei Zhang and Mingchuan Zhang and Y. K. Li and Y. Wu and Daya Guo},
      year={2024},
      eprint={2402.03300},
      archivePrefix={arXiv},
      primaryClass={cs.CL},
      url={https://arxiv.org/abs/2402.03300}, 
}

@inproceedings{
opd1,
title={On-Policy Distillation of Language Models: Learning from Self-Generated Mistakes},
author={Rishabh Agarwal and Nino Vieillard and Yongchao Zhou and Piotr Stanczyk and Sabela Ramos Garea and Matthieu Geist and Olivier Bachem},
booktitle={The Twelfth International Conference on Learning Representations},
year={2024},
url={https://openreview.net/forum?id=3zKtaqxLhW}
}

@article{opd2,
  author = {Kevin Lu and Thinking Machines Lab},
  title = {On-Policy Distillation},
  journal = {Thinking Machines Lab: Connectionism},
  year = {2025},
  note = {https://thinkingmachines.ai/blog/on-policy-distillation},
  doi = {10.64434/tml.20251026},
}

@inproceedings{
opd3,
title={Mini{LLM}: Knowledge Distillation of Large Language Models},
author={Yuxian Gu and Li Dong and Furu Wei and Minlie Huang},
booktitle={The Twelfth International Conference on Learning Representations},
year={2024},
url={https://openreview.net/forum?id=5h0qf7IBZZ}
}

@inproceedings{
opd4,
title={Disti{LLM}-2: A Contrastive Approach Boosts the Distillation of {LLM}s},
author={Jongwoo Ko and Tianyi Chen and Sungnyun Kim and Tianyu Ding and Luming Liang and Ilya Zharkov and Se-Young Yun},
booktitle={Forty-second International Conference on Machine Learning},
year={2025},
url={https://openreview.net/forum?id=rc65N9xIrY}
}

@misc{gspo,
      title={Group Sequence Policy Optimization}, 
      author={Chujie Zheng and Shixuan Liu and Mingze Li and Xiong-Hui Chen and Bowen Yu and Chang Gao and Kai Dang and Yuqiong Liu and Rui Men and An Yang and Jingren Zhou and Junyang Lin},
      year={2025},
      eprint={2507.18071},
      archivePrefix={arXiv},
      primaryClass={cs.LG},
      url={https://arxiv.org/abs/2507.18071}, 
}

@misc{flowrl,
      title={FlowRL: Matching Reward Distributions for LLM Reasoning}, 
      author={Xuekai Zhu and Daixuan Cheng and Dinghuai Zhang and Hengli Li and Kaiyan Zhang and Che Jiang and Youbang Sun and Ermo Hua and Yuxin Zuo and Xingtai Lv and Qizheng Zhang and Lin Chen and Fanghao Shao and Bo Xue and Yunchong Song and Zhenjie Yang and Ganqu Cui and Ning Ding and Jianfeng Gao and Xiaodong Liu and Bowen Zhou and Hongyuan Mei and Zhouhan Lin},
      year={2025},
      eprint={2509.15207},
      archivePrefix={arXiv},
      primaryClass={cs.LG},
      url={https://arxiv.org/abs/2509.15207}, 
}

@misc{whyopsdfail,
      title={Why Does Self-Distillation (Sometimes) Degrade the Reasoning Capability of LLMs?}, 
      author={Jeonghye Kim and Xufang Luo and Minbeom Kim and Sangmook Lee and Dohyung Kim and Jiwon Jeon and Dongsheng Li and Yuqing Yang},
      year={2026},
      eprint={2603.24472},
      archivePrefix={arXiv},
      primaryClass={cs.CL},
      url={https://arxiv.org/abs/2603.24472}, 
}

@misc{ding2026hdpo,
      title={HDPO: Hybrid Distillation Policy Optimization via Privileged Self-Distillation}, 
      author={Ken Ding},
      year={2026},
      eprint={2603.23871},
      archivePrefix={arXiv},
      primaryClass={cs.LG},
      url={https://arxiv.org/abs/2603.23871}, 
}

@inproceedings{
Speculativeopd,
title={Speculative Knowledge Distillation: Bridging the Teacher-Student Gap Through Interleaved Sampling},
author={Wenda Xu and Rujun Han and Zifeng Wang and Long Le and Dhruv Madeka and Lei Li and William Yang Wang and Rishabh Agarwal and Chen-Yu Lee and Tomas Pfister},
booktitle={The Thirteenth International Conference on Learning Representations},
year={2025},
url={https://openreview.net/forum?id=EgJhwYR2tB}
}

@misc{sdzero,
      title={Self-Distillation Zero: Self-Revision Turns Binary Rewards into Dense Supervision}, 
      author={Yinghui He and Simran Kaur and Adithya Bhaskar and Yongjin Yang and Jiarui Liu and Narutatsu Ri and Liam Fowl and Abhishek Panigrahi and Danqi Chen and Sanjeev Arora},
      year={2026},
      eprint={2604.12002},
      archivePrefix={arXiv},
      primaryClass={cs.CL},
      url={https://arxiv.org/abs/2604.12002}, 
}

@inproceedings{luffy,
 author = {Yan, Jianhao and Li, Yafu and Hu, Zican and Wang, Zhi and Cui, Ganqu and Qu, Xiaoye and Cheng, Yu and Zhang, Yue},
 booktitle = {Advances in Neural Information Processing Systems},
 editor = {D. Belgrave and C. Zhang and H. Lin and R. Pascanu and P. Koniusz and M. Ghassemi and N. Chen},
 pages = {117157--117186},
 publisher = {Curran Associates, Inc.},
 title = {Learning to Reason under Off-Policy Guidance},
 url = {https://proceedings.neurips.cc/paper_files/paper/2025/file/a9d5c33e38442450f9781bd7bad3c81e-Paper-Conference.pdf},
 volume = {38},
 year = {2025}
}

@inproceedings{aligndistill,
    title = "{A}lign{D}istil: Token-Level Language Model Alignment as Adaptive Policy Distillation",
    author = "Zhang, Songming  and
      Zhang, Xue  and
      Zhang, Tong  and
      Hu, Bojie  and
      Chen, Yufeng  and
      Xu, Jinan",
    editor = "Che, Wanxiang  and
      Nabende, Joyce  and
      Shutova, Ekaterina  and
      Pilehvar, Mohammad Taher",
    booktitle = "Proceedings of the 63rd Annual Meeting of the Association for Computational Linguistics (Volume 1: Long Papers)",
    month = jul,
    year = "2025",
    address = "Vienna, Austria",
    publisher = "Association for Computational Linguistics",
    url = "https://aclanthology.org/2025.acl-long.972/",
    doi = "10.18653/v1/2025.acl-long.972",
    pages = "19791--19807",
    ISBN = "979-8-89176-251-0"
}

@inproceedings{
tsd-kd,
title={Explain in Your Own Words: Improving Reasoning via Token-Selective Dual Knowledge Distillation},
author={Minsang Kim and Seung Jun Baek},
booktitle={The Fourteenth International Conference on Learning Representations},
year={2026},
url={https://openreview.net/forum?id=zph7e5JaXc}
}

@misc{aime24,
      title={American Invitational Mathematics Examination (AIME) 2024}, 
      author={Zhang, Yifan and Math-AI, Team},
      year={2024},
}

@misc{aime25,
      title={American Invitational Mathematics Examination (AIME) 2025}, 
      author={Zhang, Yifan and Math-AI, Team},
      year={2025},
}

@inproceedings{
openthought,
title={OpenThoughts: Data Recipes for Reasoning Models},
author={Etash Kumar Guha and Ryan Marten and Sedrick Keh and Negin Raoof and Georgios Smyrnis and Hritik Bansal and Marianna Nezhurina and Jean Mercat and Trung Vu and Zayne Rea Sprague and Ashima Suvarna and Benjamin Feuer and Leon Liangyu Chen and Zaid Khan and Eric Frankel and Sachin Grover and Caroline Choi and Niklas Muennighoff and Shiye Su and Wanjia Zhao and John Yang and Shreyas Pimpalgaonkar and Kartik sharma and Charlie Cheng-Jie Ji and Yichuan Deng and Sarah M Pratt and Vivek Ramanujan and Jon Saad-Falcon and Jeffrey Li and Achal Dave and Alon Albalak and Kushal Arora and Blake Wulfe and Chinmay Hegde and Greg Durrett and Sewoong Oh and Mohit Bansal and Saadia Gabriel and Aditya Grover and Kai-Wei Chang and Vaishaal Shankar and Aaron Gokaslan and Mike A Merrill and Tatsunori Hashimoto and Yejin Choi and Jenia Jitsev and Reinhard Heckel and Maheswaran Sathiamoorthy and Alex Dimakis and Ludwig Schmidt},
booktitle={First Workshop on Foundations of Reasoning in Language Models},
year={2025},
url={https://openreview.net/forum?id=mbqvBA12Dx}
}

@misc{LoRA,
      title={LoRA: Low-Rank Adaptation of Large Language Models}, 
      author={Edward J. Hu and Yelong Shen and Phillip Wallis and Zeyuan Allen-Zhu and Yuanzhi Li and Shean Wang and Lu Wang and Weizhu Chen},
      year={2021},
      eprint={2106.09685},
      archivePrefix={arXiv},
      primaryClass={cs.CL},
      url={https://arxiv.org/abs/2106.09685}, 
}

@inproceedings{word2vec,
  title={Efficient Estimation of Word Representations in Vector Space},
  author={Tomas Mikolov and Kai Chen and Gregory S. Corrado and Jeffrey Dean},
  booktitle={International Conference on Learning Representations},
  year={2013},
  url={https://api.semanticscholar.org/CorpusID:5959482}
}

@inproceedings{learn2Steer,
 author = {Parekh, Jayneel and KHAYATAN, Pegah and Shukor, Mustafa and Dapogny, Arnaud and Newson, Alasdair and Cord, Matthieu},
 booktitle = {Advances in Neural Information Processing Systems},
 editor = {D. Belgrave and C. Zhang and H. Lin and R. Pascanu and P. Koniusz and M. Ghassemi and N. Chen},
 pages = {159799--159834},
 publisher = {Curran Associates, Inc.},
 title = {Learning to Steer: Input-dependent Steering for Multimodal LLMs},
 url = {https://proceedings.neurips.cc/paper_files/paper/2025/file/ea491e2d1c46686b8db5cd11154f5d2c-Paper-Conference.pdf},
 volume = {38},
 year = {2025}
}

@misc{steerllm,
      title={Steering Language Models With Activation Engineering}, 
      author={Alexander Matt Turner and Lisa Thiergart and Gavin Leech and David Udell and Juan J. Vazquez and Ulisse Mini and Monte MacDiarmid},
      year={2024},
      eprint={2308.10248},
      archivePrefix={arXiv},
      primaryClass={cs.CL},
      url={https://arxiv.org/abs/2308.10248}, 
}

@inproceedings{steerLlama,
    title = "Steering Llama 2 via Contrastive Activation Addition",
    author = "Rimsky, Nina  and
      Gabrieli, Nick  and
      Schulz, Julian  and
      Tong, Meg  and
      Hubinger, Evan  and
      Turner, Alexander",
    editor = "Ku, Lun-Wei  and
      Martins, Andre  and
      Srikumar, Vivek",
    booktitle = "Proceedings of the 62nd Annual Meeting of the Association for Computational Linguistics (Volume 1: Long Papers)",
    month = aug,
    year = "2024",
    address = "Bangkok, Thailand",
    publisher = "Association for Computational Linguistics",
    url = "https://aclanthology.org/2024.acl-long.828/",
    doi = "10.18653/v1/2024.acl-long.828",
    pages = "15504--15522", 
}

@misc{refusalllm,
      title={Refusal in Language Models Is Mediated by a Single Direction}, 
      author={Andy Arditi and Oscar Obeso and Aaquib Syed and Daniel Paleka and Nina Panickssery and Wes Gurnee and Neel Nanda},
      year={2024},
      eprint={2406.11717},
      archivePrefix={arXiv},
      primaryClass={cs.LG},
      url={https://arxiv.org/abs/2406.11717}, 
}

@inproceedings{
zhao2025mitigating,
title={Mitigating Object Hallucination in Large Vision-Language Models via Image-Grounded Guidance},
author={Linxi Zhao and Yihe Deng and Weitong Zhang and Quanquan Gu},
booktitle={Forty-second International Conference on Machine Learning},
year={2025},
url={https://openreview.net/forum?id=w0xYx9CJhY}
}

@misc{blackobxopd,
      title={Black-Box On-Policy Distillation of Large Language Models}, 
      author={Tianzhu Ye and Li Dong and Zewen Chi and Xun Wu and Shaohan Huang and Furu Wei},
      year={2026},
      eprint={2511.10643},
      archivePrefix={arXiv},
      primaryClass={cs.CL},
      url={https://arxiv.org/abs/2511.10643}, 
}

@inproceedings{
piflow,
title={pi-Flow: Policy-Based Few-Step Generation via Imitation Distillation},
author={Hansheng Chen and Kai Zhang and Hao Tan and Leonidas Guibas and Gordon Wetzstein and Sai Bi},
booktitle={The Fourteenth International Conference on Learning Representations},
year={2026},
url={https://openreview.net/forum?id=1vAte8dsap}
}

@inproceedings{refinedpolicydistillation,
    title={{Refined Policy Distillation}: {F}rom {VLA} Generalists to {RL} Experts}, 
    author={Tobias Jülg and Wolfram Burgard and Florian Walter},
    year={2025},
    booktitle={Proc.~of the IEEE/RSJ Int.~Conf.~on Intelligent Robots and Systems (IROS)}
}

@misc{vllm,
      title={Efficient Memory Management for Large Language Model Serving with PagedAttention}, 
      author={Woosuk Kwon and Zhuohan Li and Siyuan Zhuang and Ying Sheng and Lianmin Zheng and Cody Hao Yu and Joseph E. Gonzalez and Hao Zhang and Ion Stoica},
      year={2023},
      eprint={2309.06180},
      archivePrefix={arXiv},
      primaryClass={cs.LG},
      url={https://arxiv.org/abs/2309.06180}, 
}

@software{trl,
  title   = {{TRL: Transformers Reinforcement Learning}},
  author  = {von Werra, Leandro and Belkada, Younes and Tunstall, Lewis and Beeching, Edward and Thrush, Tristan and Lambert, Nathan and Huang, Shengyi and Rasul, Kashif and Gallouédec, Quentin},
  license = {Apache-2.0},
  url     = {https://github.com/huggingface/trl},
  year    = {2020}
}

@inproceedings{reflection,
 author = {Shinn, Noah and Cassano, Federico and Gopinath, Ashwin and Narasimhan, Karthik and Yao, Shunyu},
 booktitle = {Advances in Neural Information Processing Systems},
 editor = {A. Oh and T. Naumann and A. Globerson and K. Saenko and M. Hardt and S. Levine},
 pages = {8634--8652},
 publisher = {Curran Associates, Inc.},
 title = {Reflexion: language agents with verbal reinforcement learning},
 url = {https://proceedings.neurips.cc/paper_files/paper/2023/file/1b44b878bb782e6954cd888628510e90-Paper-Conference.pdf},
 volume = {36},
 year = {2023}
}

@inproceedings{llmjudge,
 author = {Zheng, Lianmin and Chiang, Wei-Lin and Sheng, Ying and Zhuang, Siyuan and Wu, Zhanghao and Zhuang, Yonghao and Lin, Zi and Li, Zhuohan and Li, Dacheng and Xing, Eric and Zhang, Hao and Gonzalez, Joseph and Stoica, Ion},
 booktitle = {Advances in Neural Information Processing Systems},
 editor = {A. Oh and T. Naumann and A. Globerson and K. Saenko and M. Hardt and S. Levine},
 pages = {46595--46623},
 publisher = {Curran Associates, Inc.},
 title = {Judging LLM-as-a-Judge with MT-Bench and Chatbot Arena},
 url = {https://proceedings.neurips.cc/paper_files/paper/2023/file/91f18a1287b398d378ef22505bf41832-Paper-Datasets_and_Benchmarks.pdf},
 volume = {36},
 year = {2023}
}

@misc{maa_amc,
	title        = {American Mathematics Competitions},
	author       = {{Mathematical Association of America}},
	year         = 2025,
	note         = {Accessed: 2025-09-13}
}

@inproceedings{hendrycks2021math,
	title        = {Measuring Mathematical Problem Solving With the {MATH} Dataset},
	author       = {Hendrycks, Dan and Burns, Collin and Kadavath, Saurav and Arora, Akul and Basart, Steven and Tang, Eric and Song, Dawn and Steinhardt, Jacob},
	year         = 2021,
	booktitle    = {NeurIPS Datasets and Benchmarks Track}
}

@article{hmmt,
      title={Beyond Benchmarks: MathArena as an Evaluation Platform for Mathematics with LLMs}, 
      author={Jasper Dekoninck and Nikola Jovanović and Tim Gehrunger and Kári Rögnvaldsson and Ivo Petrov and Chenhao Sun and Martin Vechev},
      year={2026},
      eprint={2605.00674},
      archivePrefix={arXiv},
      primaryClass={cs.CL},
      url={https://arxiv.org/abs/2605.00674}, 
}

@misc{gpt5,
      title={OpenAI GPT-5 System Card}, 
      author={Aaditya Singh and Adam Fry and Adam Perelman and Adam Tart and Adi Ganesh and Ahmed El-Kishky and Aidan McLaughlin and Aiden Low and AJ Ostrow and Akhila Ananthram and Akshay Nathan and Alan Luo and Alec Helyar and Aleksander Madry and Aleksandr Efremov and Aleksandra Spyra and Alex Baker-Whitcomb and Alex Beutel and Alex Karpenko and Alex Makelov and Alex Neitz and Alex Wei and Alexandra Barr and Alexandre Kirchmeyer and Alexey Ivanov and Alexi Christakis and Alistair Gillespie and Allison Tam and Ally Bennett and Alvin Wan and Alyssa Huang and Amy McDonald Sandjideh and Amy Yang and Ananya Kumar and Andre Saraiva and Andrea Vallone and Andrei Gheorghe and Andres Garcia Garcia and Andrew Braunstein and Andrew Liu and Andrew Schmidt and Andrey Mereskin and Andrey Mishchenko and Andy Applebaum and Andy Rogerson and Ann Rajan and Annie Wei and Anoop Kotha and Anubha Srivastava and Anushree Agrawal and Arun Vijayvergiya and Ashley Tyra and Ashvin Nair and Avi Nayak and Ben Eggers and Bessie Ji and Beth Hoover and Bill Chen and Blair Chen and Boaz Barak and Borys Minaiev and Botao Hao and Bowen Baker and Brad Lightcap and Brandon McKinzie and Brandon Wang and Brendan Quinn and Brian Fioca and Brian Hsu and Brian Yang and Brian Yu and Brian Zhang and Brittany Brenner and Callie Riggins Zetino and Cameron Raymond and Camillo Lugaresi and Carolina Paz and Cary Hudson and Cedric Whitney and Chak Li and Charles Chen and Charlotte Cole and Chelsea Voss and Chen Ding and Chen Shen and Chengdu Huang and Chris Colby and Chris Hallacy and Chris Koch and Chris Lu and Christina Kaplan and Christina Kim and CJ Minott-Henriques and Cliff Frey and Cody Yu and Coley Czarnecki and Colin Reid and Colin Wei and Cory Decareaux and Cristina Scheau and Cyril Zhang and Cyrus Forbes and Da Tang and Dakota Goldberg and Dan Roberts and Dana Palmie and Daniel Kappler and Daniel Levine and Daniel Wright and Dave Leo and David Lin and David Robinson and Declan Grabb and Derek Chen and Derek Lim and Derek Salama and Dibya Bhattacharjee and Dimitris Tsipras and Dinghua Li and Dingli Yu and DJ Strouse and Drew Williams and Dylan Hunn and Ed Bayes and Edwin Arbus and Ekin Akyurek and Elaine Ya Le and Elana Widmann and Eli Yani and Elizabeth Proehl and Enis Sert and Enoch Cheung and Eri Schwartz and Eric Han and Eric Jiang and Eric Mitchell and Eric Sigler and Eric Wallace and Erik Ritter and Erin Kavanaugh and Evan Mays and Evgenii Nikishin and Fangyuan Li and Felipe Petroski Such and Filipe de Avila Belbute Peres and Filippo Raso and Florent Bekerman and Foivos Tsimpourlas and Fotis Chantzis and Francis Song and Francis Zhang and Gaby Raila and Garrett McGrath and Gary Briggs and Gary Yang and Giambattista Parascandolo and Gildas Chabot and Grace Kim and Grace Zhao and Gregory Valiant and Guillaume Leclerc and Hadi Salman and Hanson Wang and Hao Sheng and Haoming Jiang and Haoyu Wang and Haozhun Jin and Harshit Sikchi and Heather Schmidt and Henry Aspegren and Honglin Chen and Huida Qiu and Hunter Lightman and Ian Covert and Ian Kivlichan and Ian Silber and Ian Sohl and Ibrahim Hammoud and Ignasi Clavera and Ikai Lan and Ilge Akkaya and Ilya Kostrikov and Irina Kofman and Isak Etinger and Ishaan Singal and Jackie Hehir and Jacob Huh and Jacqueline Pan and Jake Wilczynski and Jakub Pachocki and James Lee and James Quinn and Jamie Kiros and Janvi Kalra and Jasmyn Samaroo and Jason Wang and Jason Wolfe and Jay Chen and Jay Wang and Jean Harb and Jeffrey Han and Jeffrey Wang and Jennifer Zhao and Jeremy Chen and Jerene Yang and Jerry Tworek and Jesse Chand and Jessica Landon and Jessica Liang and Ji Lin and Jiancheng Liu and Jianfeng Wang and Jie Tang and Jihan Yin and Joanne Jang and Joel Morris and Joey Flynn and Johannes Ferstad and Johannes Heidecke and John Fishbein and John Hallman and Jonah Grant and Jonathan Chien and Jonathan Gordon and Jongsoo Park and Jordan Liss and Jos Kraaijeveld and Joseph Guay and Joseph Mo and Josh Lawson and Josh McGrath and Joshua Vendrow and Joy Jiao and Julian Lee and Julie Steele and Julie Wang and Junhua Mao and Kai Chen and Kai Hayashi and Kai Xiao and Kamyar Salahi and Kan Wu and Karan Sekhri and Karan Sharma and Karan Singhal and Karen Li and Kenny Nguyen and Keren Gu-Lemberg and Kevin King and Kevin Liu and Kevin Stone and Kevin Yu and Kristen Ying and Kristian Georgiev and Kristie Lim and Kushal Tirumala and Kyle Miller and Lama Ahmad and Larry Lv and Laura Clare and Laurance Fauconnet and Lauren Itow and Lauren Yang and Laurentia Romaniuk and Leah Anise and Lee Byron and Leher Pathak and Leon Maksin and Leyan Lo and Leyton Ho and Li Jing and Liang Wu and Liang Xiong and Lien Mamitsuka and Lin Yang and Lindsay McCallum and Lindsey Held and Liz Bourgeois and Logan Engstrom and Lorenz Kuhn and Louis Feuvrier and Lu Zhang and Lucas Switzer and Lukas Kondraciuk and Lukasz Kaiser and Manas Joglekar and Mandeep Singh and Mandip Shah and Manuka Stratta and Marcus Williams and Mark Chen and Mark Sun and Marselus Cayton and Martin Li and Marvin Zhang and Marwan Aljubeh and Matt Nichols and Matthew Haines and Max Schwarzer and Mayank Gupta and Meghan Shah and Melody Y. Guan and Melody Huang and Meng Dong and Mengqing Wang and Mia Glaese and Micah Carroll and Michael Lampe and Michael Malek and Michael Sharman and Michael Zhang and Michele Wang and Michelle Pokrass and Mihai Florian and Mikhail Pavlov and Miles Wang and Ming Chen and Mingxuan Wang and Minnia Feng and Mo Bavarian and Molly Lin and Moose Abdool and Mostafa Rohaninejad and Nacho Soto and Natalie Staudacher and Natan LaFontaine and Nathan Marwell and Nelson Liu and Nick Preston and Nick Turley and Nicklas Ansman and Nicole Blades and Nikil Pancha and Nikita Mikhaylin and Niko Felix and Nikunj Handa and Nishant Rai and Nitish Keskar and Noam Brown and Ofir Nachum and Oleg Boiko and Oleg Murk and Olivia Watkins and Oona Gleeson and Pamela Mishkin and Patryk Lesiewicz and Paul Baltescu and Pavel Belov and Peter Zhokhov and Philip Pronin and Phillip Guo and Phoebe Thacker and Qi Liu and Qiming Yuan and Qinghua Liu and Rachel Dias and Rachel Puckett and Rahul Arora and Ravi Teja Mullapudi and Raz Gaon and Reah Miyara and Rennie Song and Rishabh Aggarwal and RJ Marsan and Robel Yemiru and Robert Xiong and Rohan Kshirsagar and Rohan Nuttall and Roman Tsiupa and Ronen Eldan and Rose Wang and Roshan James and Roy Ziv and Rui Shu and Ruslan Nigmatullin and Saachi Jain and Saam Talaie and Sam Altman and Sam Arnesen and Sam Toizer and Sam Toyer and Samuel Miserendino and Sandhini Agarwal and Sarah Yoo and Savannah Heon and Scott Ethersmith and Sean Grove and Sean Taylor and Sebastien Bubeck and Sever Banesiu and Shaokyi Amdo and Shengjia Zhao and Sherwin Wu and Shibani Santurkar and Shiyu Zhao and Shraman Ray Chaudhuri and Shreyas Krishnaswamy and Shuaiqi and Xia and Shuyang Cheng and Shyamal Anadkat and Simón Posada Fishman and Simon Tobin and Siyuan Fu and Somay Jain and Song Mei and Sonya Egoian and Spencer Kim and Spug Golden and SQ Mah and Steph Lin and Stephen Imm and Steve Sharpe and Steve Yadlowsky and Sulman Choudhry and Sungwon Eum and Suvansh Sanjeev and Tabarak Khan and Tal Stramer and Tao Wang and Tao Xin and Tarun Gogineni and Taya Christianson and Ted Sanders and Tejal Patwardhan and Thomas Degry and Thomas Shadwell and Tianfu Fu and Tianshi Gao and Timur Garipov and Tina Sriskandarajah and Toki Sherbakov and Tomek Korbak and Tomer Kaftan and Tomo Hiratsuka and Tongzhou Wang and Tony Song and Tony Zhao and Troy Peterson and Val Kharitonov and Victoria Chernova and Vineet Kosaraju and Vishal Kuo and Vitchyr Pong and Vivek Verma and Vlad Petrov and Wanning Jiang and Weixing Zhang and Wenda Zhou and Wenlei Xie and Wenting Zhan and Wes McCabe and Will DePue and Will Ellsworth and Wulfie Bain and Wyatt Thompson and Xiangning Chen and Xiangyu Qi and Xin Xiang and Xinwei Shi and Yann Dubois and Yaodong Yu and Yara Khakbaz and Yifan Wu and Yilei Qian and Yin Tat Lee and Yinbo Chen and Yizhen Zhang and Yizhong Xiong and Yonglong Tian and Young Cha and Yu Bai and Yu Yang and Yuan Yuan and Yuanzhi Li and Yufeng Zhang and Yuguang Yang and Yujia Jin and Yun Jiang and Yunyun Wang and Yushi Wang and Yutian Liu and Zach Stubenvoll and Zehao Dou and Zheng Wu and Zhigang Wang},
      year={2026},
      eprint={2601.03267},
      archivePrefix={arXiv},
      primaryClass={cs.CL},
      url={https://arxiv.org/abs/2601.03267}, 
}

@inproceedings{sanchez2024stay,
  title={Stay on Topic with Classifier-Free Guidance},
  author={Sanchez, Guillaume and Spangher, Alexander and Fan, Honglu and Levi, Elad and Biderman, Stella},
  booktitle={International Conference on Machine Learning},
  pages={43197--43234},
  year={2024},
  organization={PMLR}
}
